\documentclass[preprint,5p]{elsarticle}

\usepackage{epsfig}
\usepackage{amsthm}

\usepackage{lineno}

\usepackage{microtype}
\usepackage[utf8]{inputenc}
\usepackage[english]{babel}
\usepackage{graphicx}
\usepackage{subfigure}
\usepackage{caption}
\usepackage{algorithm}
\usepackage{multirow}
\usepackage{makecell}
\usepackage[table,xcdraw]{xcolor}
\usepackage{subfiles}
\usepackage{blindtext}
\usepackage{verbatim}
\usepackage{booktabs} 
\usepackage{amsmath,amssymb,amsfonts}
\usepackage{algorithmic}
\usepackage{graphicx}  
\usepackage{textcomp}
\usepackage{xcolor}
\usepackage{csquotes}
 \usepackage{amssymb}
\usepackage{hyperref}
\usepackage{xltabular,longtable,ltxtable}
\newcommand{\tabincell}[2]{\begin{tabular}{@{}#1@{}}#2\end{tabular}}

\usepackage{fancyhdr}
\usepackage{lipsum}
\journal{Knowledge-Based Systems}

\begin{document}

\begin{frontmatter}



\title{AutoML: A Survey of the State-of-the-Art}


\author{Xin He} \ead{csxinhe@comp.hkbu.edu.hk}
\author{Kaiyong Zhao} \ead{kyzhao@comp.hkbu.edu.hk}
\author{Xiaowen Chu\corref{cor}} \ead{chxw@comp.hkbu.edu.hk}

\address{Department of Computer Science, Hong Kong Baptist University}
\cortext[cor]{Corresponding author}

\begin{abstract}

Deep learning (DL) techniques have obtained remarkable achievements on various tasks, such as image recognition, object detection, and language modeling. However, building a high-quality DL system for a specific task highly relies on human expertise, hindering its wide application. Meanwhile, automated machine learning (AutoML) is a promising solution for building a DL system without human assistance and is being extensively studied. This paper presents a comprehensive and up-to-date review of the state-of-the-art (SOTA) in AutoML. According to the DL pipeline, we introduce AutoML methods –– covering data preparation, feature engineering, hyperparameter optimization, and neural architecture search (NAS) –– with a particular focus on NAS, as it is currently a hot sub-topic of AutoML. We summarize the representative NAS algorithms' performance on the CIFAR-10 and ImageNet datasets and further discuss the following subjects of NAS methods: one/two-stage NAS, one-shot NAS, joint hyperparameter and architecture optimization, and resource-aware NAS. Finally, we discuss some open problems related to the existing AutoML methods for future research.

\end{abstract}


\end{frontmatter}



\section{Introduction}\label{section:introduction}

In recent years, deep learning has been applied in various fields and used to solve many challenging AI tasks, in areas such as image classification \cite{alexnet, resnet}, object detection \cite{yolo}, and language modeling \cite{frage, transformerxl}. Specifically, since AlexNet \cite{alexnet} outperformed all other traditional manual methods in the 2012 ImageNet Large Scale Visual Recognition Challenge (ILSVRC) \cite{ILSVRC15}, increasingly complex and deep neural networks have been proposed. For example, VGG-16 \cite{vggnet} has more than 130 million parameters, occupies nearly 500 MB of memory space, and requires 15.3 billion floating-point operations to process an image of size $224\times224$. Notably, however, these models were all manually designed by experts by a trial-and-error process, which means that even experts require substantial resources and time to create well-performing models.

\begin{figure*}[ht]
    \centering
    \includegraphics[width=\textwidth]{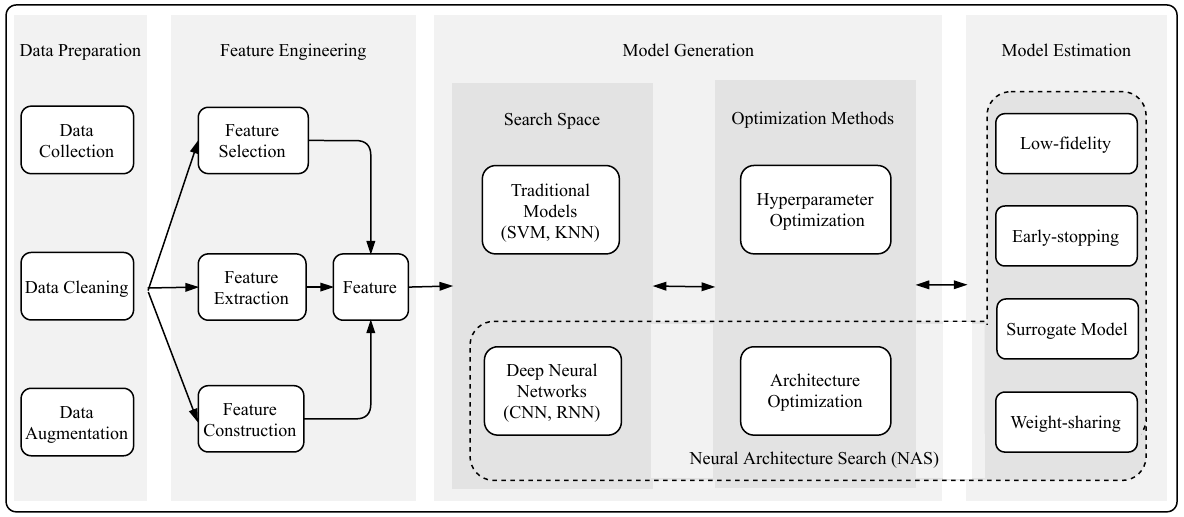}
    \caption{An overview of AutoML pipeline covering data preparation (Section \ref{section:data_preparation}), feature engineering (Section \ref{section:feature_engineering}), model generation (Section \ref{section:model_generation}) and model evaluation (Section \ref{section:model_estimation}). }
    \label{fig:pipeline_details}
\end{figure*}

To reduce these onerous development costs, a novel idea of automating the entire pipeline of machine learning (ML) has emerged, i.e., automated machine learning (AutoML). There are various definitions of AutoML. For example, according to \cite{automl_benchmark}, AutoML is designed to reduce the demand for data scientists and enable domain experts to automatically build ML applications without much requirement for statistical and ML knowledge. In \cite{takeout_automl_survey}, AutoML is defined as a combination of automation and ML. In a word, AutoML can be understood to involve the automated construction of an ML pipeline on the limited computational budget. With the exponential growth of computing power, AutoML has become a hot topic in both industry and academia. A complete AutoML system can make a dynamic combination of various techniques to form an easy-to-use end-to-end ML pipeline system (as shown in Figure \ref{fig:pipeline_details}). Many AI companies have created and publicly shared such systems (e.g., Cloud AutoML  \footnote{https://cloud.google.com/automl/} by Google) to help people with little or no ML knowledge to build high-quality custom models.

As Figure \ref{fig:pipeline_details} shows, the AutoML pipeline consists of several processes: data preparation, feature engineering, model generation, and model evaluation. Model generation can be further divided into \textit{search space} and \textit{optimization methods}. The \textit{search space} defines the design principles of ML models, which can be divided into two categories: the traditional ML models (e.g., SVM and KNN), and neural architectures. The optimization methods are classified into \textit{hyperparameter optimization (HPO)} and \textit{architecture optimization (AO)}, where the former indicates the training-related parameters (e.g., the learning rate and batch size), and the latter indicates the model-related parameters (e.g., the number of layer for neural architectures and the number of neighbors for KNN). NAS consists of three important components: the search space of neural architectures, AO methods, and model evaluation methods. AO methods may also refer to \textit{search strategy} \cite{nas_survey} or \textit{search policy} \cite{eval_nas_phase}. Zoph et al. \cite{nas_rl_zoph16} were one of the first to propose NAS, where a recurrent network is trained by reinforcement learning to automatically search for the best-performing architecture. Since \cite{nas_rl_zoph16} successfully discovered a neural network achieving comparable results to human-designed models, there has been an explosion of research interest in AutoML, with most focusing on NAS. NAS aims to search for a robust and well-performing neural architecture by selecting and combining different basic operations from a predefined search space. By reviewing NAS methods, we classify the commonly used search space into \textit{entire-structured} \cite{nas_rl_zoph16,enas,smash_onehot}, \textit{cell-based} \cite{enas,nasnet_zoph17,blockqnn,darts_liu18,pnas_liu18}, \textit{hierarchical} \cite{liu_hierarchical_2018} and \textit{morphism-based} \cite{chen2015net2net,network_morphism_wei16,autokeras} search space. The commonly used AO methods contain \textit{reinforcement learning} (RL) \cite{nas_rl_zoph16,nasnet_zoph17,metaqnn,blockqnn,enas}, \textit{evolution-based algorithm} (EA) \cite{stanley_evolving_2002,large_scale_evolve,amoebanet,lemonade,cgp_suganuma17,evolve_dnn,genertic_cnn}, and \textit{gradient descent} (GD) \cite{darts_liu18,maskconnect_Ahmed18,shin_differentiable_nas_2018}, \textit{Surrogate Model-Based Optimization} (SMBO) \cite{towards_autotune,towards_nas_hpo,fabolas,practical_hpo_Falkner18,smbo_random_forest,bohb,bergstra_making_nodate}, and hybrid AO methods \cite{cars,evolv_hyrid_agent,reinf_evolv,surrogate_evolv,diff_evolv}.


Although there are already several excellent AutoML-related surveys \cite{nas_survey,a_nas_survey,comprehensive_nas,takeout_automl_survey,automl_benchmark}, to the best of our knowledge, our survey covers a broader range of AutoML methods. As summarized in Table \ref{table:survey_compare}, \cite{nas_survey,a_nas_survey,comprehensive_nas} only focus on NAS, while \cite{takeout_automl_survey,automl_benchmark} cover little of NAS technique. In this paper, we summarize the AutoML-related methods according to the complete AutoML pipeline (Figure \ref{fig:pipeline_details}), providing beginners with a comprehensive introduction to the field. Notably, many sub-topics of AutoML are large enough to have their own surveys. However, our goal is not to conduct a thorough investigation of all AutoML sub-topics. Instead, we focus on the breadth of research in the field of AutoML. Therefore, we will summarize and discuss some representative methods of each process in the pipeline. 


\begin{table}[!ht]
    \centering
    \begin{tabular}{c|c|c|c|c}\hline
        \textbf{Survey} & \textbf{DP} & \textbf{FE}& \textbf{HPO} & \textbf{NAS} \\\hline
        
        NAS Survey \cite{nas_survey}&-&-&-&\checkmark\\
        A Survey on NAS \cite{a_nas_survey}&-&-&-&\checkmark\\
        NAS Challenges \cite{comprehensive_nas}&-&-&-&\checkmark\\
        A Survey on AutoML \cite{takeout_automl_survey}&-&\checkmark&\checkmark&\dag\\
        AutoML Challenges \cite{automl_challenge}&\checkmark&-&\checkmark&\dag\\
        AutoML Benchmark \cite{automl_benchmark}&\checkmark&\checkmark&\checkmark&-\\
        Ours&\checkmark&\checkmark&\checkmark&\checkmark\\
        \hline
    \end{tabular}
    \caption{Comparison between different AutoML surveys. The ``Survey'' column gives each survey a label based on their title for increasing the readability. \textit{DP, FE, HPO, NAS} indicate data preparation, feature engineering, hyperparameter optimization and neural architecture search, respectively. ``-'', ``\checkmark'', and ``\dag'' indicate the content is 1) not mentioned; 2) mentioned detailed; 3) mentioned briefly, in the original paper, respectively.}
    \label{table:survey_compare}
\end{table}

The rest of this paper is organized as follows. The processes of data preparation, feature engineering, model generation, and model evaluation are presented in Sections \ref{section:data_preparation}, \ref{section:feature_engineering}, \ref{section:model_generation}, \ref{section:model_estimation}, respectively. In Section \ref{section:nas_performance}, we compare the performance of NAS algorithms on the CIFAR-10 and ImageNet dataset, and discuss several subtopics of great concern in NAS community: one/two-stage NAS, one-shot NAS, joint hyperparameter and architecture optimization, and resource-aware NAS. In Section \ref{section:open_problems}, we describe several open problems in AutoML. We conclude our survey in Section \ref{section:conclusion}.

\section{Data Preparation}\label{section:data_preparation}

\begin{figure}[!ht]
    \centering
    \includegraphics[width=0.48\textwidth]{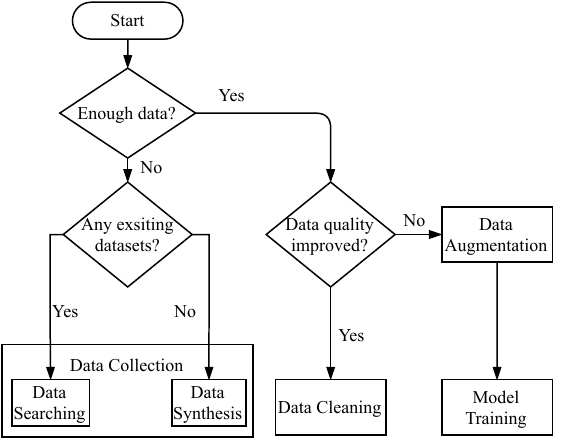}
    \caption{The flow chart for data preparation.}
    \label{fig:data_preparation}
\end{figure}

The first step in the ML pipeline is data preparation. Figure \ref{fig:data_preparation} presents the workflow of data preparation, which can be introduced in three aspects: data collection, data cleaning, and data augmentation. Data collection is a necessary step to build a new dataset or extend the existing dataset. The process of data cleaning is used to filter noisy data so that downstream model training is not compromised. Data augmentation plays an important role in enhancing model robustness and improving model performance. The following subsections will cover the three aspects in more detail.

\subsection{Data Collection}

ML's deepening study has led to a consensus that high-quality datasets are of critical importance for ML; as a result, numerous open datasets have emerged. In the early stages of ML study, a handwritten digital dataset, i.e., MNIST \cite{mnist}, was developed. After that, several larger datasets like CIFAR-10 and CIFAR-100 \cite{cifar10} and ImageNet \cite{imagenet} were developed. A variety of datasets can also be retrieved by entering the keywords into these websites: Kaggle \footnote{https://www.kaggle.com}, Google Dataset Search (GOODS) \footnote{https://datasetsearch.research.google.com/}, and Elsevier Data Search \footnote{https://www.datasearch.elsevier.com/}. 

However, it is usually challenging to find a proper dataset through the above approaches for some particular tasks, such as those related to medical care or other privacy matters. Two types of methods are proposed to solve this problem: data searching and data synthesis. 

\subsubsection{Data Searching}

As the Internet is an inexhaustible data source, searching for Web data is an intuitive way to collect a dataset \cite{filter_web_data,chen2013neil,xia2014well,do2015automatic}. However, there are some problems with using Web data. 

First, the search results may not exactly match the keywords. Thus, unrelated data must be filtered. For example, Krause et al. \cite{krause2016unreasonable} separate inaccurate results as cross-domain or cross-category noise, and remove any images that appear in search results for more than one category. Vo et al. \cite{vo2017harnessing} re-rank relevant results and provide search results linearly, according to keywords. 

Second, Web data may be incorrectly labeled or even unlabeled. A learning-based self-labeling method is often used to solve this problem. For example, the active learning method \cite{collins2008towards} selects the most ‘‘uncertain'' unlabeled individual examples for labeling by a human, and then iteratively labels the remaining data. Roh et al. \cite{survey_data_collect} provided a review of semi-supervised learning self-labeling methods, which can help take the human out of the loop of labeling to improve efficiency, and can be divided into the following categories: self-training \cite{yarowsky1995unsupervised, triguero2014characterization}, co-training \cite{hady2010combining, blum1998combining}, and co-learning \cite{democratic_co_learn}. Moreover, due to the complexity of Web images content, a single label cannot adequately describe an image. Consequently, Yang et al. \cite{filter_web_data} assigned multiple labels to a Web image, i.e., if the confidence scores of these labels are very close or the label with the highest score is the same as the original label of the image, then this image will be set as a new training sample.

However, the distribution of Web data can be extremely different from that of the target dataset, which will increase the difficulty of training the model. A common solution is to fine-tune these Web data \cite{chen2015webly,xu2015augmenting}. Yang et al. \cite{filter_web_data} proposed an iterative algorithm for model training and Web data-filtering. Dataset imbalance is another common problem, as some special classes have a very limited number of Web data. To solve this problem, the synthetic minority over-sampling technique (SMOTE) \cite{chawla2002smote} is used to synthesize new minority samples between existing real minority samples, instead of simply up-sampling minority samples or down-sampling the majority samples. In another approach, Guo et al. \cite{guo2004learning} combined the boosting method with data generation to enhance the generalizability and robustness of the model against imbalanced data sets.

\subsubsection{Data Synthesis}

Data simulator is one of the most commonly used methods to generate data. For some particular tasks, such as autonomous driving, it is not possible to test and adjust a model in the real world during the research phase, due to safety hazards. Therefore, a practical approach to generating data is to use a data simulator that matches the real world as closely as possible. OpenAI Gym \cite{openai_gym} is a popular toolkit that provides various simulation environments, in which developers can concentrate on designing their algorithms, instead of struggling to generate data. Wang et al. \cite{irs} used a popular game engine, Unreal Engine 4, to build a large synthetic indoor robotics stereo (IRS) dataset, which provides the information for disparity and surface normal estimation. Furthermore, a reinforcement learning-based method is applied in \cite{ruiz2018learning} for optimizing the parameters of a data simulator to control the distribution of the synthesized data.


Another novel technique for deriving synthetic data is \textit{Generative Adversarial Networks} (GANs) \cite{goodfellow2014generative}, which can be used to generate images \cite{goodfellow2014generative,oh2018learning,rendergan,gan_aug}, tabular \cite{table_gan, tabular_gan} and text \cite{text_gan_no_RL} data. Karras et al. \cite{karras2018style} applied GAN technique to generate realistic human face images. Oh and Jaroensri et al. \cite{oh2018learning} built a synthetic dataset, which captures small motion for video-motion magnification. Bowles et al. \cite{gan_aug} demonstrated the feasibility of using GAN to generate medical images for brain segmentation tasks. In the case of textual data, applying GAN to text has proved difficult because the commonly used method is to use reinforcement learning to update the gradient of the generator, but the text is discrete, and thus the gradient cannot propagate from discriminator to generator. To solve this problem, Donahue et al. \cite{text_gan_no_RL} used an autoencoder to encode sentences into a smooth sentence representation to remove the barrier of reinforcement learning. Park et al. \cite{table_gan} applied GAN to synthesize fake tables that are statistically similar to the original table but do not cause information leakage. Similarly, in \cite{tabular_gan}, GAN is applied to generate tabular data like medical or educational records.


\subsection{Data Cleaning}

The collected data inevitably have noise, but the noise can negatively affect the training of the model. Therefore, the process of data cleaning \cite{dc_survey, jesmeen2018survey} must be carried out if necessary. Across the literature, the effort of data cleaning is shifting from crowdsourcing to automation. Traditionally, data cleaning requires specialist knowledge, but access to specialists is limited and generally expensive. Hence, Chu et al. \cite{katara} proposed Katara, a knowledge-based and crowd-powered data cleaning system. To improve efficiency, some studies \cite{sampleclean, activeclean} proposed only to clean a small subset of the data and maintain comparable results to the case of cleaning the full dataset. However, these methods require a data scientist to design what data cleaning operations are applied to the dataset. BoostClean \cite{boostclean} attempts to automate this process by treating it as a boosting problem. Each data cleaning operation effectively adds a new cleaning operation to the input of the downstream ML model, and through a combination of Boosting and feature selection, a good series of cleaning operations, which can well improve the performance of the ML model, can be generated. AlphaClean \cite{alphaclean} transforms data cleaning into a hyperparameter optimization problem, which further increases automation. Specifically, the final data cleaning combinatorial operation in AlphaClean is composed of several pipelined cleaning operations that need to be searched from a predefined search space. Gemp et al. \cite{auto_dc} attempted to use meta-learning technique to automate the process of data cleaning.

The data cleaning methods mentioned above are applied to a fixed dataset. However, the real world generates vast amounts of data every day. In other words, how to clean data in a continuous process becomes a worth studying problem, especially for enterprises. Ilyas et al. \cite{eval_continuous_dc} proposed an effective way of evaluating the algorithms of continuously cleaning data. Mahdavi et al. \cite{auto_dc_workflow} built a cleaning workflow orchestrator, which can learn from previous cleaning tasks, and proposed promising cleaning workflows for new datasets.

\subsection{Data Augmentation}

\begin{figure}
    \centering
    \includegraphics[width=0.48\textwidth]{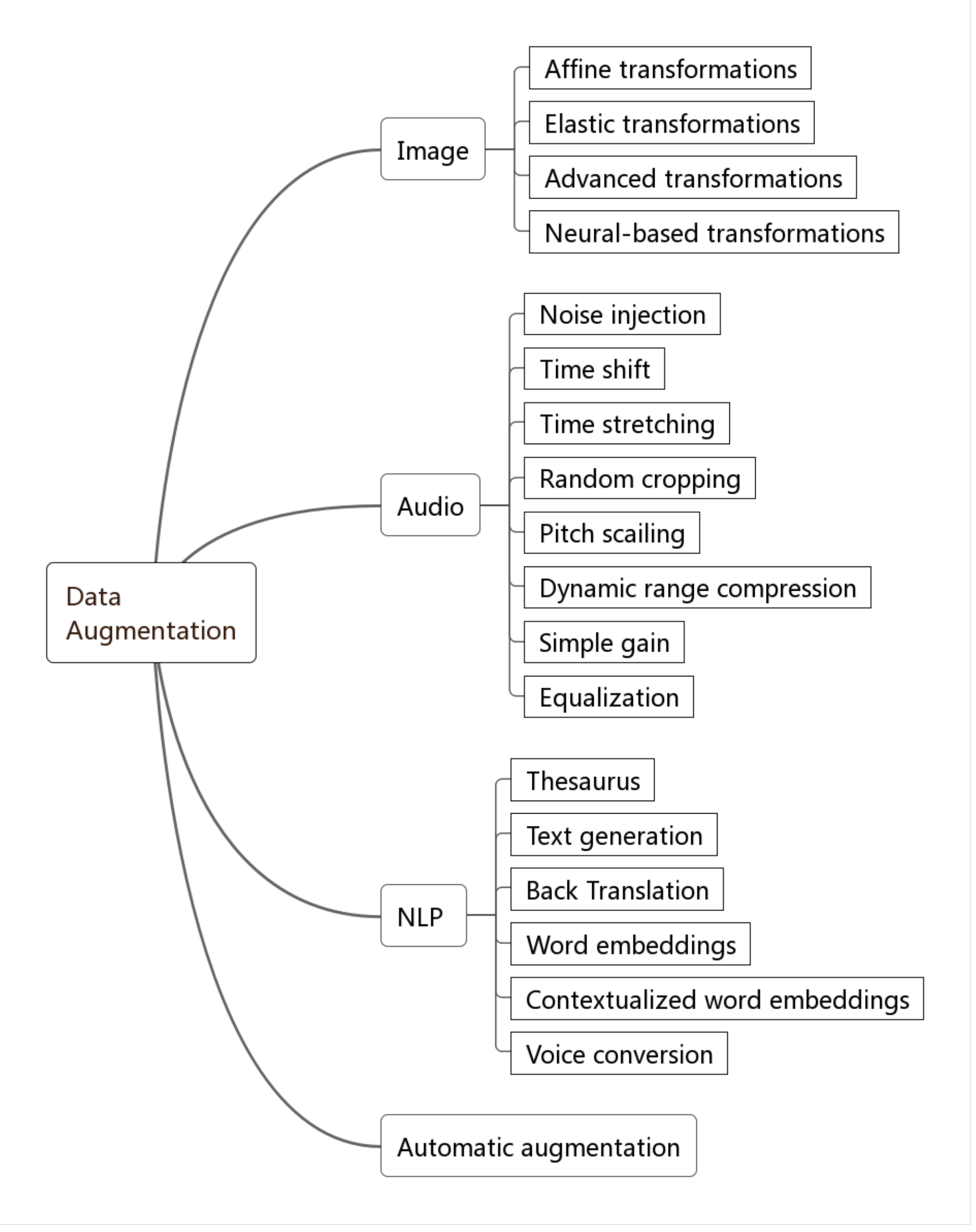}
    \caption{A classification of data augmentation techniques.}
    \label{fig:data_augmentation}
\end{figure}

To some degree, data augmentation (DA) can also be regarded as a tool for data collection, as it can generate new data based on the existing data. However, DA also serves as a regularizer to avoid over-fitting of model training and has received more and more attention. Therefore, we introduce DA as a separate part of data preparation in detail. Figure \ref{fig:data_augmentation} classifies DA techniques from the perspective of data type (image, audio, and text), and incorporates automatic DA techniques that have recently received much attention. 

For image data, the affine transformations include rotation, scaling, random cropping, and reflection; the elastic transformations contain the operations like contrast shift, brightness shift, blurring, and channel shuffle; the advanced transformations involve random erasing, image blending, cutout \cite{cutout}, and mixup \cite{mixup}, etc. These three types of common transformations are available in some open source libraries, like torchvision \footnote{https://pytorch.org/docs/stable/torchvision/transforms.html}, ImageAug \cite{imgaug}, and Albumentations \cite{albumentations}. In terms of neural-based transformations, it can be divided into three categories: adversarial noise \cite{aug_an}, neural style transfer \cite{st_aug}, and GAN technique \cite{aug_gan}. For textual data, Wong et al. \cite{wong2016understanding} proposed two approaches for creating additional training examples: data warping and synthetic over-sampling. The former generates additional samples by applying transformations to data-space, and the latter creates additional samples in feature-space. Textual data can be augmented by synonym insertion or by first translating the text into a foreign language and then translating it back to the original language. In a recent study, Xie et al. \cite{xie2017data} proposed a non-domain-specific DA policy that uses noising in RNNs, and this approach works well for the tasks of language modeling and machine translation. Yu et al. \cite{yu2018qanet} proposed a back-translation method for DA to improve reading comprehension. NLPAug \cite{nlp_aug} is an open-source library that integrates many types of augmentation operations for both textual and audio data.

The above augmentation techniques still require human to select augmentation operations and then form a specific DA policy for specific tasks, which requires much expertise and time. Recently, there are many methods \cite{cubuk2019autoaugment,dada,fasteraa,fastaa,greedy_aa,population_aa,aug_policy,rs_aug,adv_aug,online_aug,uniform_aug} proposed to search for augmentation policy for different tasks. AutoAugment \cite{cubuk2019autoaugment} is a pioneering work to automate the search for optimal DA policies using reinforcement learning. However, AutoAugment is not efficient as it takes almost 500 GPU hours for one augmentation search. In order to improve search efficiency, a number of improved algorithms have subsequently been proposed using different search strategies, such as gradient descent-based \cite{dada,fasteraa}, Bayesian-based optimization \cite{fastaa}, online hyperparameter learning \cite{online_aug}, greedy-based search \cite{greedy_aa} and random search \cite{rs_aug}. Besides, LingChen et al. \cite{uniform_aug} proposed a search-free DA method, namely UniformAugment, by assuming that the augmentation space is approximately distribution invariant.

\section{Feature Engineering}\label{section:feature_engineering}
It is generally accepted that data and features determine the upper bound of ML, and that models and algorithms can only approximate this limit. In this context, feature engineering aims to maximize the extraction of features from raw data for use by algorithms and models. Feature engineering consists of three sub-topics: feature selection, feature extraction, and feature construction. Feature extraction and construction are variants of feature transformation, by which a new set of features is created \cite{feat_SelExtCon_motoda02}. In most cases, feature extraction aims to reduce the dimensionality of features by applying specific mapping functions, while feature construction is used to expand original feature spaces, and the purpose of feature selection is to reduce feature redundancy by selecting important features. Thus, the essence of automatic feature engineering is, to some degree, a dynamic combination of these three processes. 

\subsection{Feature Selection}

Feature selection builds a feature subset based on the original feature set by reducing irrelevant or redundant features. This tends to simplify the model, hence avoiding overfitting and improving model performance. The selected features are usually divergent and highly correlated with object values. According to \cite{dash1997feature}, there are four basic steps in a typical process of feature selection (see Figure \ref{fig:fe_process}), as follows:

\begin{figure}[ht]
    \centering
    \includegraphics[width=0.3\textwidth]{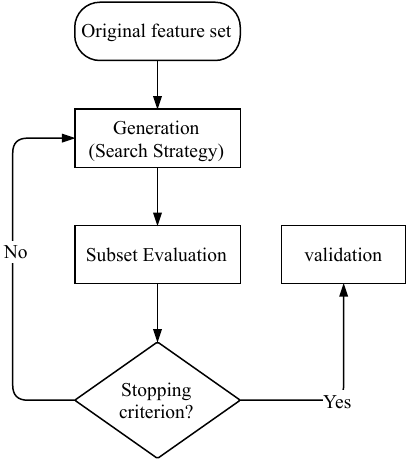}
    \caption{The iterative process of feature selection. A subset of features is selected, based on a search strategy, and then evaluated. Then, a validation procedure is implemented to determine whether the subset is valid. The above steps are repeated until the stop criterion is satisfied.}
    \label{fig:fe_process}
\end{figure}

The search strategy for feature selection involves three types of algorithms: complete search, heuristic search, and random search. Complete search comprises exhaustive and non-exhaustive searching; the latter can be further split into four methods: breadth-first search, branch and bound search, beam search, and best-first search. Heuristic search comprises sequential forward selection (SFS), sequential backward selection (SBS), and bidirectional search (BS). In SFS and SBS, the features are added from an empty set or removed from a full set, respectively, whereas BS uses both SFS and SBS to search until these two algorithms obtain the same subset. The most commonly used random search methods are simulated annealing (SA) and genetic algorithms (GAs).

Methods of subset evaluation can be divided into three different categories. The first is the filter method, which scores each feature according to its divergence or correlation and then selects features according to a threshold. Commonly used scoring criteria for each feature are variance, the correlation coefficient, the chi-square test, and mutual information. The second is the wrapper method, which classifies the sample set with the selected feature subset, after which the classification accuracy is used as the criterion to measure the quality of the feature subset. The third method is the embedded method, in which variable selection is performed as part of the learning procedure. Regularization, decision tree, and deep learning are all embedded methods.

%

\subsection{Feature Construction}

Feature construction is a process that constructs new features from the basic feature space or raw data to enhance the robustness and generalizability of the model. Essentially, this is done to increase the representative ability of the original features. This process is traditionally highly dependent on human expertise, and one of the most commonly used methods is preprocessing transformation, such as standardization, normalization, or feature discretization. In addition, the transformation operations for different types of features may vary. For example, operations such as conjunctions, disjunctions and negation are typically used for Boolean features; operations such as minimum, maximum, addition, subtraction, mean are typically used for numerical features, and operations such as Cartesian product \cite{pazzani1998constructive} and M-of-N \cite{zheng1998comparison} are commonly used for nominal features. 

It is impossible to manually explore all possibilities. Hence, to further improve efficiency, some automatic feature construction methods \cite{gama2004functional,zheng1998comparison,vafaie1998evolutionary,feat_cons_survey} have been proposed to automate the process of searching and evaluating the operation combination, and shown to achieve results as good as or superior to those achieved by human expertise. Besides, some feature construction methods, such as decision tree-based methods \cite{gama2004functional,zheng1998comparison} and genetic algorithms \cite{vafaie1998evolutionary}, require a predefined operation space, while the annotation-based approaches \cite{feat_cons_survey} do not, as they can  use domain knowledge (in the form of annotation) and the training examples, and hence, can be traced back to the interactive feature-space construction protocol introduced by \cite{roth2009interactive}. Using this protocol, the learner identifies inadequate regions of feature space and, in coordination with a domain expert, adds descriptiveness using existing semantic resources. After selecting possible operations and constructing a new feature, feature-selection techniques are applied to evaluate the new feature.


\subsection{Feature Extraction}

Feature extraction is a dimensionality-reduction process performed via some mapping functions. It extracts informative and non-redundant features according to certain metrics. Unlike feature selection, feature extraction alters the original features. The kernel of feature extraction is a mapping function, which can be implemented in many ways. The most prominent approaches are principal component analysis (PCA), independent component analysis, isomap, nonlinear dimensionality reduction, and linear discriminant analysis (LDA). Recently, the feed-forward neural network approach has become popular; this uses the hidden units of a pretrained model as extracted features. Furthermore, many autoencoder-based algorithms are proposed; for example, Zeng et al. \cite{meng2017relational} proposed a relation autoencoder model that considers data features and their relationships, while an unsupervised feature-extraction method using autoencoder trees is proposed by \cite{irsoy2017unsupervised}.


\section{Model Generation}\label{section:model_generation}

Model generation is divided into two parts––search space and optimization methods––as shown in Figure \ref{fig:pipeline_details}. The search space defines the model structures that can be designed and optimized in principle. The types of models can be broadly divided into two categories: traditional ML models, such as support vector machine (SVM) \cite{svm} and k-nearest neighbors algorithm (KNN) \cite{KNN}, and deep neural network (DNN). There are two types of parameters for the optimization methods: hyperparameters used for training, such as the learning rate, and those used for model design, such as the filter size and the number of layers for DNN. Neural architecture search (NAS) has recently attracted considerable attention; therefore, in this section, we introduce the search space and optimization methods of NAS technique. Readers who are interested in traditional models (e.g., SVM) can refer to other reviews \cite{takeout_automl_survey,automl_benchmark}.

\begin{figure}
    \centering
    \includegraphics[width=0.48\textwidth]{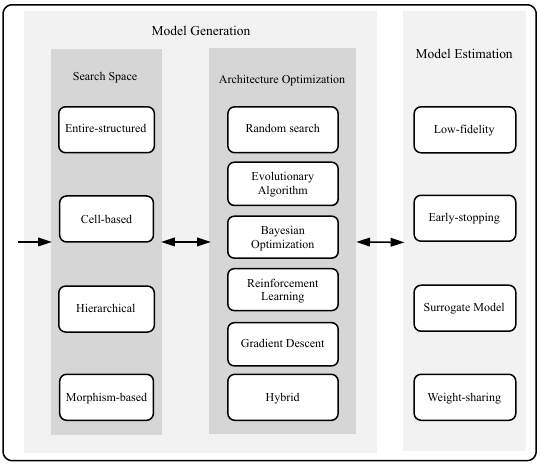}
    \caption{An overview of neural architecture search pipeline.}
    \label{fig:nas_pipeline}
\end{figure}

Figure \ref{fig:nas_pipeline} presents an overview of the NAS pipeline, which is categorized into the following three dimensions \cite{nas_survey,nas_eval}: search space, architecture optimization (AO) method\footnote{It can also be referred to as the ``search strategy \cite{nas_survey,nas_eval}", ``search policy \cite{eval_nas_phase}", or ``optimization method \cite{a_nas_survey,takeout_automl_survey}".}, and model evaluation method. 

\begin{itemize}
    \item \textit{Search Space}. The search space defines the design principles of neural architectures. Different scenarios require different search spaces. Here, we summarize four types of commonly used search spaces: entire-structured, cell-based, hierarchical, and morphism-based.
    \item \textit{Architecture Optimization Method}. The architecture optimization (AO) method defines how to guide the search to efficiently find the model architecture with high performance after the search space is defined.
    \item \textit{Model Evaluation Method}. Once a model is generated, its performance needs to be evaluated. The simplest approach of doing this is to train the model to converge on the training set, and then estimate model performance on the validation set; however, this method is time-consuming and resource-intensive. Some advanced methods can accelerate the evaluation process but lose fidelity in the process. Thus, how to balance the efficiency and effectiveness of an evaluation is a problem worth studying.
\end{itemize}

The search space and AO methods are presented in this section, while the methods of model evaluation are presented in the next section.


\subsection{Search Space}

A neural architecture can be represented as a direct acyclic graph (DAG) comprising $B$ ordered nodes. In DAG, each node and directed edge indicate a feature tensor and an operation, respectively. Eq. \ref{eq:ss} presents a formula for computation at any node $Z_k, k\in\{1,2,...,B\}$.


\begin{figure}[ht]
    \centering
    \includegraphics[width=0.36\textwidth]{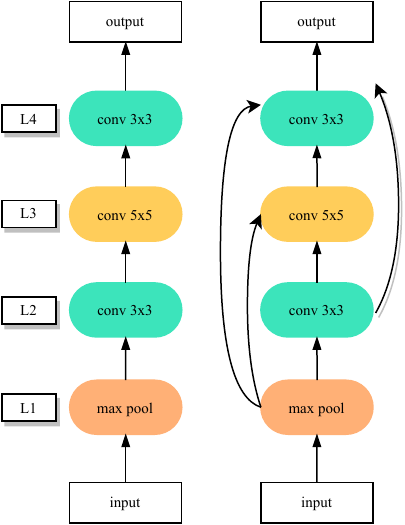}
    \caption{Two simplified examples of entire-structured neural architectures. Each layer is specified with a different operation, such as convolution and max-pooling operations. The edge indicates the information flow. The skip-connection operation used in the right example can help explore deeper and more complex neural architectures.}
    \label{fig:whole_structure}
\end{figure}

\begin{equation}
    Z_k = \sum_{i=1}^{N_k} o_i(I_i), \,\, o_i\in O
    \label{eq:ss}
\end{equation}

\noindent where $N_k$ indicates the indegree of node $Z_k$, $I_i$ and $o_i$ represent $i$-th input tensor and its associated operation, respectively, and $O$ is a set of candidate operations, such as convolution, pooling, activation functions, skip connection, concatenation, and addition. To further enhance the model performance, many NAS methods use certain advanced human-designed modules as primitive operations, such as depth-wise separable convolution \cite{chollet2017xception}, dilated convolution\cite{dilated_conv}, and squeeze-and-excitation (SE) blocks \cite{senet}. The selection and combination of these operations vary with the design of search space. In other words, the search space defines the structural paradigm that AO methods can explore; thus, designing a good search space is a vital but challenging problem. In general, a good search space is expected to exclude human bias and be flexible enough to cover a wider variety of model architectures. Based on the existing NAS studies, we detail the commonly used search spaces as follows.

\subsubsection{Entire-structured Search Space} 

The space of entire-structured neural networks \cite{nas_rl_zoph16,enas} is one of the most intuitive and straightforward search spaces. Figure \ref{fig:whole_structure} presents two simplified examples of entire-structured models, which are built by stacking a predefined number of nodes, where each node represents a layer and performs a specified operation. The left model shown in Figure \ref{fig:whole_structure} indicates the simplest structure, while the right model is relatively complex, as it permits arbitrary skip connections \cite{resnet} to exist between the ordered nodes; these connections have been proven effective in practice \cite{nas_rl_zoph16}. Although an entire structure is easy to implement, it has several disadvantages. For example, it is widely accepted that the deeper is the model, the better is its generalization ability; however, searching for such a deep network is onerous and computationally expensive. Furthermore, the generated architecture lacks transferability; that is, a model generated on a small dataset may not fit a larger dataset, which necessitates the generation of a new model for a larger dataset.

\subsubsection{Cell-based Search Space}

\begin{figure}[!ht]
    \centering
    \includegraphics[width=0.48\textwidth]{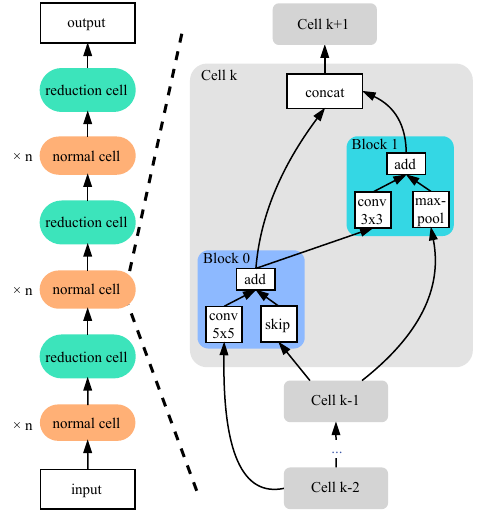}
    \caption{(Left) Example of a cell-based model comprising three motifs, each with $n$ normal cells and one reduction cell. (Right) Example of a normal cell, which contains two blocks, each having two nodes. Each node is specified with a different operation and input.}
    \label{fig:cell_structure}
\end{figure}

\textbf{Motivation}. To enable the transferability of the generated model, the cell-based search space has been proposed \cite{nasnet_zoph17,blockqnn,enas}, in which the neural architecture is composed of a fixed number of repeating cell structures. This design approach is based on the observation that many well-performing human-designed models \cite{resnet,densenet} are also built by stacking a fixed number of modules. For example, the ResNet family builds many variants, such as ResNet50, ResNet101, and ResNet152, by stacking several \textit{BottleNeck} modules \cite{resnet}. Throughout the literature, this repeated module is referred to as a motif, cell, or block, while in this paper, we call it a \textit{cell}.


\textbf{Design}. Figure \ref{fig:cell_structure} (left) presents an example of a final cell-based neural network, which comprises two types of cells: \textit{normal} and \textit{reduction} cells. Thus, the problem of searching for a full neural architecture is simplified into searching for an optimal cell structure in the context of cell-based search space. Besides, the output of the normal cell retains the same spatial dimension as the input, and the number of normal cell repeats is usually set manually based on the actual demand. The reduction cell follows behind a normal cell and has a similar structure to that of the normal cell, with the differences being that the width and height of the output feature maps of the reduction cell are half the input, and the number of channels is twice the input. This design approach follows the common practice of manually designing neural networks. Unlike the entire-structured search space, the model built on cell-based search space can be expanded to form a larger model by simply adding more cells without re-searching for the cell structure. Meanwhile, many approaches \cite{darts_liu18, enas, nasnet_zoph17} have experimentally demonstrated the transferability of the model generated in cell-based search space, such as the model built on CIFAR-10, which can also achieve comparable results to those achieved by SOTA human-designed models on ImageNet.

The design paradigm of the internal cell structure of most NAS studies refers to Zoph et al. \cite{nasnet_zoph17}, who were among the first to propose the exploration of cell-based search space. Figure \ref{fig:cell_structure} (right) shows an example of a normal cell structure. Each cell contains $B$ blocks (here $B=2$), and each block has two nodes. Each node in a block can be assigned different operations and receive different inputs. The output of two nodes in the block can be combined through addition or concatenation operation. Therefore, each block can be represented by a five-element tuple, $(I_1,I_2,O_1,O_2,C)$, where $I_{1}, I_{2} \in \mathcal{I}_{b}$ indicate the inputs to the block, while $O_{1}, O_{2} \in \mathcal{O}$ indicate the operations applied to inputs, and $C \in \mathcal{C}$ describes how to combine $O_1$ and $O_2$. As the blocks are ordered, the set of candidate inputs $I_b$ for the nodes in block $b_k$, which contains the output of the previous two cells and the output set of all previous blocks $\{b_i, i<k\}$ of the same cell. The first two inputs of the first cell of the whole model are set to the image data by default.

In the actual implementation, certain essential details need to be noted. First, the number of channels may differ for different inputs. A commonly used solution is to apply a calibration operation on each node's input tensor to ensure that all inputs have the same number of channels. The calibration operation generally uses $1\times1$ convolution filters, such that it will not change the width and height of the input tensor, but keep the channel number of all input tensors consistent. Second, as mentioned above, the input of a node in a block can be obtained from the previous two cells or the previous blocks within the same cell; hence, the cell's output must have the same spatial resolution. To this end, if the input/output resolutions are different, the calibration operation has stride 2; otherwise, it has stride 1. Besides, all blocks have stride 1.

\textbf{Complexity}. Searching for a cell structure is more efficient than searching for an entire structure. To illustrate this, let us assume that there are $M$ predefined candidate operations, the number of layers for both entire and the cell-based structures is $L$, and the number of blocks in a cell is $B$. Then, the number of possible entire structures can be expressed as:

\begin{equation}
N_{entire}=M^L\times 2^{\frac{L\times(L-1)}{2}}
\label{eq:entire}
\end{equation}

\noindent The number of possible cells is $(M^B\times (B+2)!)^2$. However, as there are two types of cells (i.e., normal and reduction cells), the final size of the cell-based search space is calculated as

\begin{equation}
N_{cell}=(M^B\times (B+2)!)^4
\label{eq:cell}
\end{equation}

\noindent Evidently, the complexity of searching for the entire structure grows exponentially with the number of layers. For an intuitive comparison, we assign the variables in the Eqs. \ref{eq:entire} and \ref{eq:cell} the typical value in the literature, i.e., $M=5, L=10, B=3$; then $N_{entire}=3.44\times10^{20}$ is much larger than $N_{cell}=5.06\times10^{16}$.

\begin{figure}
    \centering
    \includegraphics[width=0.48\textwidth]{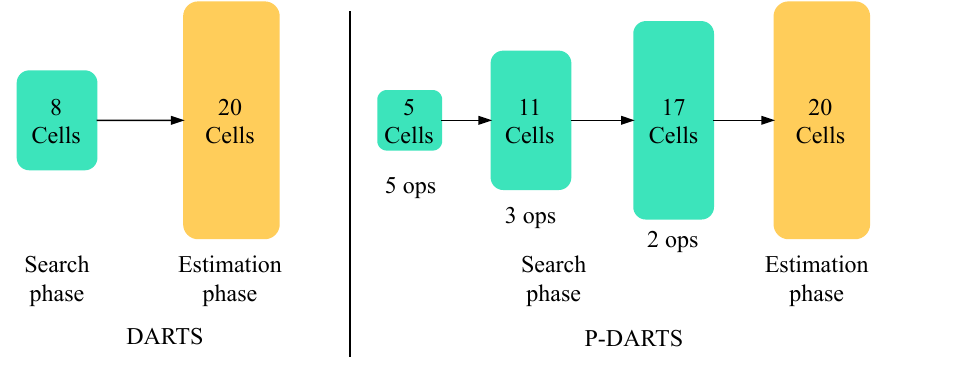}
    \caption{Difference between DARTS \cite{darts_liu18} and P-DARTS \cite{pdarts}. Both methods search and evaluate networks on the CIFAR-10 dataset. As the number of cell structures increases from 5 to 11 and then 17, the number of candidate operations is gradually reduced accordingly.}
    \label{fig:pdarts}
\end{figure}

\textbf{Two-stage Gap}. The NAS methods of cell-based search space usually comprise two phases: search and evaluation. First, in the search phase, the best-performing model is selected, and then, in the evaluation phase, it is trained from scratch or fine-tuned. However, there exists a large gap in the model depth between the two phases. As Figure \ref{fig:pdarts} (left) shows, for DARTS \cite{darts_liu18}, the generated model in the search phase only comprises eight cells for reducing the GPU memory consumption, while in the evaluation phase, the number of cells is extended to 20. Although the search phase finds the best cell structure for the shallow model, this does not mean that it is still suitable for the deeper model in the evaluation phase. In other words, simply adding more cells may deteriorate the model performance. To bridge this gap, Chen et al. \cite{pdarts} proposed an improved method based on DARTS, namely progressive-DARTS (P-DARTS), which divides the search phase into multiple stages and gradually increases the depth of the searched networks at the end of each stage, hence bridging the gap between search and evaluation. However, increasing the number of cells in the search phase may result in heavier computational overhead. Thus, for reducing the computational consumption, P-DARTS gradually reduces the number of candidate operations from 5 to 3, and then 2, through search space approximation methods, as shown in Figure \ref{fig:pdarts}. Experimentally, P-DARTS obtains a 2.50\% error rate on the CIFAR-10 test dataset, outperforming the 2.83\% error rate achieved by DARTS.

\subsubsection{Hierarchical Search Space}

 \begin{figure}[ht]
    \centering
    \includegraphics[width=0.3\textwidth]{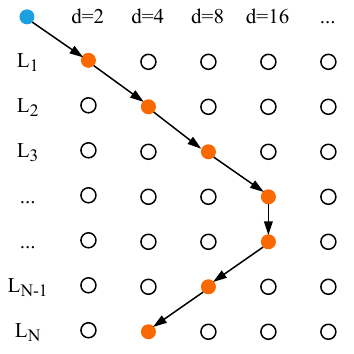}
    \caption{Network-level search space proposed by \cite{auto_deeplab}. The blue point (top-left) indicates the fixed ``stem'' structure, the remaining gray and orange points are cell structure, as described above. The black arrows along the orange points indicate the final selected network-level structure. ``d'' and ``L'' indicate the down sampling rate and layer, respectively.}
    \label{fig:two_level_hier}
\end{figure}

The cell-based search space enables the transferability of the generated model, and most of the cell-based methods \cite{enas, nasnet_zoph17, metaqnn, blockqnn, large_scale_evolve, amoebanet} follow a two-level hierarchy: the inner is the cell level, which selects the operation and connection for each node in the cell, and the outer is the network level, which controls the spatial-resolution changes. However, these approaches focus on the cell level and ignore the network level. As shown in Figure \ref{fig:cell_structure}, whenever a fixed number of normal cells are stacked, the spatial dimension of the feature maps is halved by adding a reduction cell. To jointly learn a suitable combination of repeatable cell and network structures, Liu et al. \cite{auto_deeplab} defined a general formulation for a network-level structure, depicted in Figure \ref{fig:two_level_hier}, from which many existing good network designs can be reproduced. In this way, we can fully explore the different number of channels and sizes of feature maps of each layer in the network.

\begin{figure}[ht]
    \centering
    \includegraphics[width=0.48\textwidth]{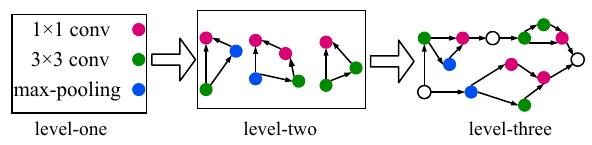}
    \caption{Example of a three-level hierarchical architecture representation. The level-one primitive operations are assembled into level-two cells. The level-two cells are viewed as primitive operations and assembled into level-three cell.}
    \label{fig:hier_cell}
\end{figure}

In terms of the cell level, the number of blocks ($B$) in a cell is still manually predefined and fixed in the search stage. In other words, $B$ is a new hyperparameter that requires tuning by human input. To address this problem, Liu et al. \cite{liu_hierarchical_2018} proposed a novel hierarchical genetic representation scheme, namely \textit{HierNAS}, in which a higher-level cell is generated by iteratively incorporating lower-level cells. As shown in Figure \ref{fig:hier_cell}, level-one cells can be some primitive operations, such as $1 \times 1$ and $3 \times 3$ convolution and $3 \times 3$ max-pooling, and are the basic components of level-two cells. Then, level-two cells are used as primitive operations to generate level-three cells. The highest-level cell is a single motif corresponding to the full architecture. Besides, a higher-level cell is defined by a learnable adjacency upper-triangular matrix $G$, where $G_{ij}=k$ indicates that the $k$-th operation $0_k$ is implemented between nodes $i$ and $j$. For example, the level-two cell shown in Figure \ref{fig:hier_cell}(a) is defined by a matrix $G$, where $G_{01}=2,G_{02}=1,G_{12}=0$ (the index starts from 0). This method can identify more types of cell structures with more complex and flexible topologies. Similarly, Liu et al. \cite{pnas_liu18} proposed \textit{progressive NAS (PNAS)} to search for the cell progressively, starting from the simplest cell structure, which is composed of only one block, and then expanding to a higher-level cell by adding more possible block structures. Moreover, PNAS improves the search efficiency by using a surrogate model to predict the top-k promising blocks from the search space at each stage of cell construction.

\begin{figure}[!ht]
    \centering
    \includegraphics[width=0.35\textwidth]{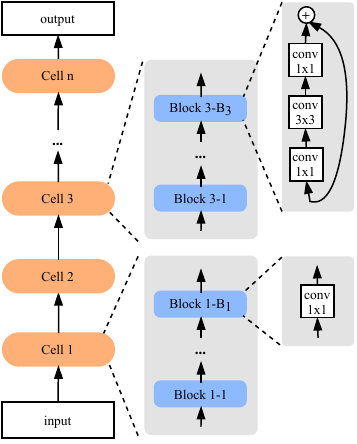}
    \caption{Factorized hierarchical search space in MnasNet \cite{mnasnet}. The final network comprises different cells. Each cell is composed of a variable number of repeated blocks, where the block in the same cell shares the same structure but differs from that in the other cells.}
    \label{fig:mnasnet}
\end{figure}

For both HierNAS and PNAS, once a cell structure is searched, it is used in all network layers, which limits the layer diversity. Besides, for achieving both high accuracy and low latency, some studies \cite{mnasnet,fbnet} proposed to search for complex and fragmented cell structures. For example, Tan et al. \cite{mnasnet} proposed MnasNet, which uses a novel factorized hierarchical search space to generate different cell structures, namely MBConv, for different layers of the final network. Figure \ref{fig:mnasnet} presents the factorized hierarchical search space of MnasNet, which comprises a predefined number of cell structures. Each cell has a different structure and contains a variable number of blocks––whereas all blocks in the same cell exhibit the same structure, those in other cells exhibit different structures. As this design method can achieve a suitable balance between model performance and latency, many subsequent studies \cite{fbnet,proxylessnas} have referred to it. Owing to the large computational consumption, most of the differentiable NAS (DNAS) techniques (e.g., DARTS) first search for a suitable cell structure on a proxy dataset (e.g., CIFAR10), and then transfer it to a larger target dataset (e.g., ImageNet). Han et al. \cite{proxylessnas} proposed ProxylessNAS, which can directly search for neural networks on the targeted dataset and hardware platforms by using BinaryConnect \cite{binaryconnect}, which addresses the high memory consumption issue.

\subsubsection{Morphism-based Search Space}

\begin{figure}[!ht]
    \centering
    \includegraphics[width=0.48\textwidth]{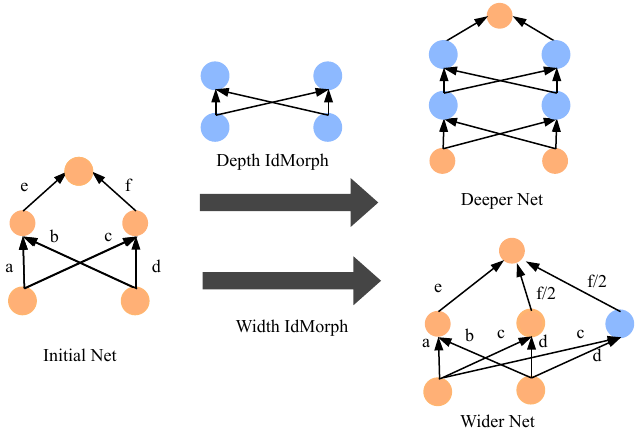}
    \caption{Net2DeeperNet and Net2WiderNet transformations in \cite{chen2015net2net}. ``IdMorph" refers to identity morphism operation. The value on each edge indicates the weight.}
    \label{fig:net2net}
\end{figure}

Isaac Newton is reported to have said that ``If I have seen further, it is by standing on the shoulders of giants.'' Similarly, several training tricks have been proposed, such as knowledge distillation \cite{knowledgedistill} and transfer learning \cite{transferlearning}. However, these methods do not directly modify the model structure. To this end, Chen et al. \cite{chen2015net2net} proposed the Net2Net technique for designing new neural networks based on an existing network by inserting identity morphism (IdMorph) transformations between the neural network layers. An IdMorph transformation is function-preserving and can be classified into two types -- depth and width IdMorph (shown in Figure \ref{fig:net2net}) -- which makes it possible to replace the original model with an equivalent model that is deeper or wider. 

However, IdMorph is limited to width and depth changes, and can only modify them separately; moreover, the sparsity of its identity layer can create problems \cite{resnet}. Therefore, an improved method is proposed, namely network morphism \cite{network_morphism_wei16}, which allows a child network to inherit all knowledge from its well-trained parent network and continue to grow into a more robust network within a shortened training time. Compared with Net2Net, network morphism exhibits the following advantages: 1) it can embed nonidentity layers and handle arbitrary nonlinear activation functions, and 2) it can simultaneously perform depth, width, and kernel size-morphing in a single operation, whereas Net2Net has to separately consider depth and width changes. The experimental results in \cite{network_morphism_wei16} show that network morphism can substantially accelerate the training process, as it uses one-fifteenth of the training time and achieves better results than the original VGG16. 

Several subsequent studies \cite{lemonade,autokeras,modularized_morph,eff_net_transform,nas_nn_mmar19,path_level_nm,fast_nm,Morphnet} are based on network morphism. For instance, Jin et al. \cite{autokeras} proposed a framework that enables Bayesian optimization to guide the network morphism for an efficient neural architecture search. Wei et al. \cite{modularized_morph} further improved network morphism at a higher level, i.e., by morphing a convolutional layer into the arbitrary module of a neural network. Additionally, Tan and Le \cite{efficientnet} proposed EfficientNet, which re-examines the effect of model scaling on convolutional neural networks, and proved that carefully balancing the network depth, width, and resolution can lead to better performance.

\subsection{Architecture Optimization}
 
After defining the search space, we need to search for the best-performing architecture, a process we call \textit{architecture optimization (AO)}. Traditionally, the architecture of a neural network is regarded as a set of static hyperparameters that are tuned based on the performance observed on the validation set. However, this process highly depends on human experts and requires considerable time and resources for trial and error. Therefore, many AO methods have been proposed to free humans from this tedious procedure and to search for novel architectures automatically. Below, we detail the commonly used AO methods.

\subsubsection{Evolutionary Algorithm}

The evolutionary algorithm (EA) is a generic population-based metaheuristic optimization algorithm that takes inspiration from biological evolution. Compared with traditional optimization algorithms such as exhaustive methods, EA is a mature global optimization method with high robustness and broad applicability. It can effectively address the complex problems that traditional optimization algorithms struggle to solve, without being limited by the problem's nature. 

\textbf{Encoding Scheme.} Different EAs may use different types of encoding schemes for network representation. There are two types of encoding schemes: direct and indirect. 

Direct encoding is a widely used method that explicitly specifies the phenotype. For example, genetic CNN \cite{genertic_cnn} encodes the network structure into a fixed-length binary string, e.g., $1$ indicates that two nodes are connected, and vice versa. Although binary encoding can be performed easily, its computational space is the square of the number of nodes, which is fixed-length, i.e., predefined manually. For representing variable-length neural networks, DAG encoding is a promising solution \cite{cgp_suganuma17,large_scale_evolve,liu_hierarchical_2018}. For example, Suganuma et al. \cite{cgp_suganuma17} used the Cartesian genetic programming (CGP) \cite{cgp_2, cgp_1} encoding scheme to represent a neural network built by a list of sub-modules that are defined as DAG. Similarly, in \cite{large_scale_evolve}, the neural architecture is also encoded as a graph, whose vertices indicate rank-3 tensors or activations (with batch normalization performed with rectified linear units (ReLUs) or plain linear units) and edges indicate identity connections or convolutions. Neuro evolution of augmenting topologies (NEAT) \cite{stanley_evolving_2002,large_scale_evolve} also uses a direct encoding scheme, where each node and connection is stored. 

Indirect encoding specifies a generation rule to build the network and allows for a more compact representation. Cellular encoding (CE) \cite{Gruau1993Cellular} is an example of a system that utilizes indirect encoding of network structures. It encodes a family of neural networks into a set of labeled trees and is based on a simple graph grammar. Some recent studies \cite{fernando2016convolution,kim2015deep,pugh2013evolving, lemonade} have described the use of indirect encoding schemes to represent a network. For example, the network in \cite{lemonade} can be encoded by a function, and each network can be modified using function-preserving network morphism operators. Hence, the child network has increased capacity and is guaranteed to perform at least as well as the parent networks.

\begin{figure}
    \centering
    \includegraphics[width=0.32\textwidth]{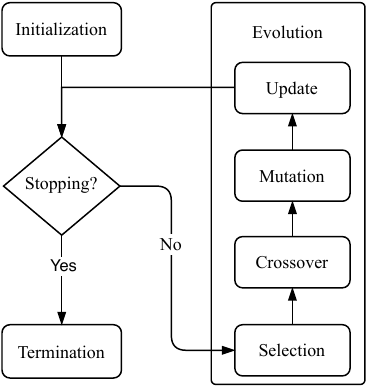}
    \caption{Overview of the evolutionary algorithm.}
    \label{fig:ea}
\end{figure}

\textbf{Four Steps.} A typical EA comprises the following steps: selection, crossover, mutation, and update (Figure \ref{fig:ea}):

\begin{itemize}
    \item \textbf{Selection} 
    This step involves selecting a portion of the networks from all generated networks for the crossover, which aims to maintain well-performing neural architectures while eliminating the weak ones. The following three strategies are adopted for network selection. The first is \textit{fitness selection}, in which the probability of a network being selected is proportional to its fitness value, i.e., $P(h_i)=\frac{Fitness(h_i)}{\sum_{j=1}^N Fitness(h_j)}$, where $h_i$ indicates the $i$-th network. The second is \textit{rank selection}, which is similar to fitness selection, but with the network’s selection probability being proportional to its relative fitness rather than its absolute fitness. The third method is \textit{tournament selection} \cite{large_scale_evolve, lemonade, amoebanet,  liu_hierarchical_2018}. Here, in each iteration, $k$ (tournament size) networks are randomly selected from the population and sorted according to their performance; then, the best network is selected with a probability of $p$, the second-best network has a probability of $p\times(1-p)$, and so on. 

    \item \textbf{Crossover}
    After selection, every two networks are selected to generate a new offspring network, inheriting half of the genetic information of each of its parents. This process is analogous to the genetic recombination, which occurs during biological reproduction and crossover. The particular manner of crossover varies and depends on the encoding scheme. In binary encoding, networks are encoded as a linear string of bits, where each bit represents a unit, such that two parent networks can be combined through one- or multiple-point crossover. However, the crossover of the data arranged in such a fashion can sometimes damage the data. Thus, Xie et al. \cite{genertic_cnn} denoted the basic unit in a crossover as a stage rather than a bit, which is a higher-level structure constructed by a binary string. For cellular encoding, a randomly selected sub-tree is cut from one parent tree to replace a sub-tree cut from the other parent tree. In another approach, NEAT performs an artificial synapsis based on historical markings, adding a new structure without losing track of the gene present throughout the simulation.

    \item \textbf{Mutation}
    As the genetic information of the parents is copied and inherited by the next generation, gene mutation also occurs. A point mutation \cite{cgp_suganuma17,genertic_cnn} is one of the most widely used operations and involves randomly and independently flipping each bit. Two types of mutations have been described in \cite{evolve_dnn}: one enables or disables a connection between two layers, and the other adds or removes skip connections between two nodes or layers. Meanwhile, Real and Moore et al. \cite{large_scale_evolve} predefined a set of mutation operators, such as altering the learning rate and removing skip connections between the nodes. By analogy with the biological process, although a mutation may appear as a mistake that causes damage to the network structure and leads to a loss of functionality, it also enables the exploration of more novel structures and ensures diversity.

    \item \textbf{Update}
    Many new networks are generated by completing the above steps, and considering the limitations on computational resources, some of these must be removed. In \cite{large_scale_evolve}, the worst-performing network of two randomly selected networks is immediately removed from the population. Alternatively, in \cite{amoebanet}, the oldest networks are removed. Other methods \cite{evolve_dnn,genertic_cnn,cgp_suganuma17} discard all models at regular intervals. However, Liu et al. \cite{liu_hierarchical_2018} did not remove any network from the population, and instead, allowed the network number to grow with time. Zhu et al. \cite{eena} regulated the population number through a variable $\lambda$, i.e., removed the worst model with probability $\lambda$ and the oldest model with $1-\lambda$.

\end{itemize}

\subsubsection{Reinforcement Learning}

\begin{figure}[ht]
    \centering
    \includegraphics[width=0.48\textwidth]{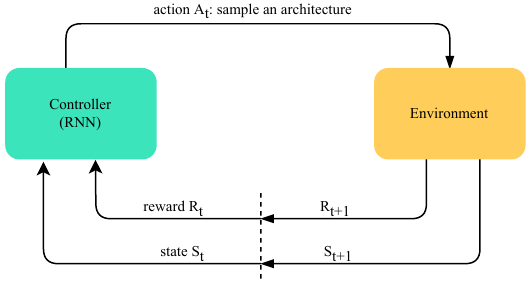}
    \caption{Overview of neural architecture search using reinforcement learning.}
    \label{fig:rl_nas}
\end{figure}

\begin{figure}[ht]
    \centering
    \includegraphics[width=0.48\textwidth]{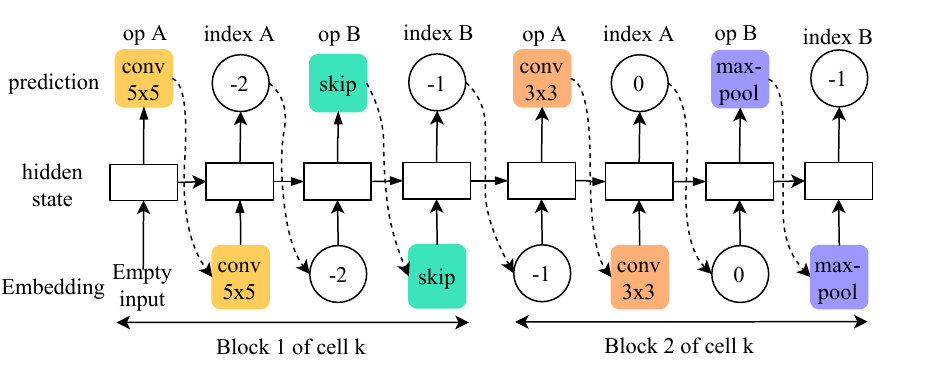}
    \caption{Example of a controller generating a cell structure. Each block in the cell comprises two nodes that are specified with different operations and inputs. The indices $-2$ and $-1$ indicate the inputs are derived from prev-previous and previous cell, respectively.}
    \label{fig:rl_cell}
\end{figure}

Zoph et al. \cite{nas_rl_zoph16} were among the first to apply reinforcement learning (RL) to neural architecture search. Figure \ref{fig:rl_nas} presents an overview of an RL-based NAS algorithm. Here, the controller is usually a recurrent neural network (RNN) that executes an action $A_t$ at each step $t$ to sample a new architecture from the search space and receives an observation of the state $S_t$ together with a reward scalar $R_t$ from the environment to update the controller's sampling strategy. Environment refers to the use of a standard neural network training procedure to train and evaluate the network generated by the controller, after which the corresponding results (such as accuracy) are returned. Many follow-up approaches \cite{metaqnn,nasnet_zoph17,blockqnn,enas} have used this framework, but with different controller policies and  neural-architecture encoding. Zoph et al. \cite{nas_rl_zoph16} first used the policy gradient algorithm \cite{policygradient} to train the controller, and sequentially sampled a string to encode the entire neural architecture. In a subsequent study \cite{nasnet_zoph17}, they used the proximal policy optimization (PPO) algorithm \cite{ppo} to update the controller, and proposed the method shown in Figure \ref{fig:rl_cell} to build a cell-based neural architecture. MetaQNN \cite{metaqnn} is a meta-modeling algorithm using Q-learning with an $\epsilon$-greedy exploration strategy and experience replay to sequentially search for neural architectures. 

Although the above RL-based algorithms have achieved SOTA results on the CIFAR-10 and Penn Treebank (PTB) \cite{ptb} datasets, they incur considerable time and computational resources. For instance, the authors in \cite{nas_rl_zoph16} took 28 days and 800 K40 GPUs to search for the best-performing architecture, and MetaQNN \cite{metaqnn} also took 10 days and 10 GPUs to complete its search. To this end, some improved RL-based algorithms have been proposed. BlockQNN \cite{blockqnn} uses a distributed asynchronous framework and an early-stop strategy to complete searching on only one GPU within 20 hours. The efficient neural architecture search (ENAS) \cite{enas} is even better, as it adopts a parameter-sharing strategy in which all child architectures are regarded as sub-graphs of a supernet; this enables these architectures to share parameters, obviating the need to train each child model from scratch. Thus, ENAS took only approximately 10 hours using one GPU to search for the best architecture on the CIFAR-10 dataset, which is nearly 1000$\times$ faster than \cite{nas_rl_zoph16}.


\begin{figure*}[!ht]
    \centering
    \includegraphics[width=0.8\textwidth]{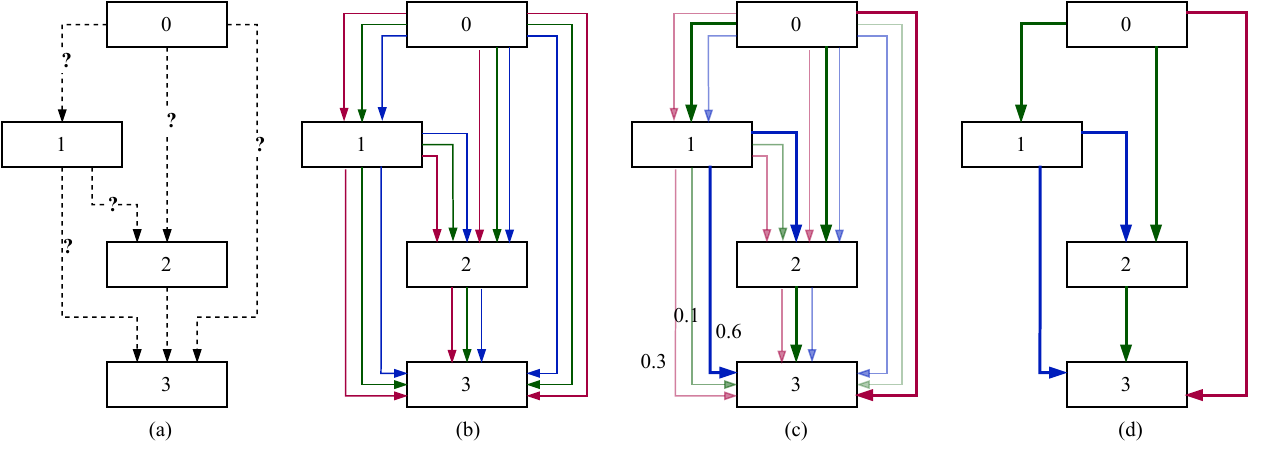}
    \caption{Overview of DARTS. (a) The data can only flow from lower-level nodes to higher-level nodes, and the operations on edges are initially unknown. (b) The initial operation on each edge is a mixture of candidate operations, each having equal weight. (c) The weight of each operation is learnable and ranges from 0 to 1, but for previous discrete sampling methods, the weight could only be 0 or 1. (d) The final neural architecture is constructed by preserving the maximum weight-value operation on each edge.}
    \label{fig:darts}
\end{figure*}

\subsubsection{Gradient Descent}

The above-mentioned search strategies sample neural architectures from a discrete search space. A pioneering algorithm, namely DARTS \cite{darts_liu18}, was among the first gradient descent (GD)-based method to search for neural architectures over a continuous and differentiable search space by using a softmax function to relax the discrete space, as outlined below:


\begin{equation}
\overline{o}_{i,j}(x)=\sum_{k=1}^{K} \frac{\exp \left(\alpha^{k}_{i,j}\right)}{\sum_{l=1}^{K} \exp \left(\alpha^{l}_{i,j}\right)} o^k(x)
\label{eq:darts_softmax}
\end{equation}

\noindent where $o(x)$ indicates the operation performed on input $x$, $\alpha^k_{i,j}$ indicates the weight assigned to the operation $o^k$ between a pair of nodes ($i,j$), and $K$ is the number of predefined candidate operations. After the relaxation, the task of searching for architectures is transformed into a joint optimization of neural architecture $\alpha$ and the weights of this neural architecture $\theta$. These two types of parameters are optimized alternately, indicating a bilevel optimization problem. Specifically, $\alpha$ and $\theta$ are optimized with the validation and the training sets, respectively. The training and the validation losses are denoted by $\mathcal{L}_{train}$ and $\mathcal{L}_{val}$, respectively. Hence, the total loss function can be derived as follows:

\begin{equation}
\begin{array}{cl}
{\min_{\alpha}} & {\mathcal{L}_{val}\left(\theta^*, \alpha\right)} \\
{\text { s.t. }} & {\theta^*=\operatorname{argmin}_{\theta} \, \mathcal{L}_{train}(\theta, \alpha)}
\end{array}
\label{eq:bilevel}
\end{equation}

Figure \ref{fig:darts} presents an overview of DARTS, where a cell is composed of $N$ (here $N=4$) ordered nodes and the node $z^k$ ($k$ starts from 0) is connected to the node $z^{i}, i\in\{k+1,...,N\}$. The operation on each edge $e_{i,j}$ is initially a mixture of candidate operations, each being of equal weight. Therefore, the neural architecture $\alpha$ is a supernet that contains all possible child neural architectures. At the end of the search, the final architecture is derived by retaining only the maximum-weight operation among all mixed operations.

Although DARTS substantially reduces the search time, it incurs several problems. First, as Eq. \ref{eq:bilevel} shows, DARTS describes a joint optimization of the neural architecture and weights as a bilevel optimization problem. However, this problem is difficult to solve directly, because both architecture $\alpha$ and weights $\theta$ are high dimensional parameters. Another solution is single-level optimization, which can be formalized as

\begin{equation}
    \min _{\theta, \alpha} \mathcal{L}_{\mathrm{train}}(\theta, \alpha)
    \label{eq:single_level}
\end{equation}

\noindent which optimizes both neural architecture and weights together. Although the single-level optimization problem can be efficiently solved as a regular training, the searched architecture $\alpha$ commonly overfits the training set and its performance on the validation set cannot be guaranteed. The authors in \cite{MiLeNAS} proposed mixed-level optimization:

\begin{equation}
\min _{\alpha, \theta}\,\,\left[\mathcal{L}_{train}\left(\theta^{*}, \alpha\right)+\lambda \mathcal{L}_{val}\left(\theta^{*}, \alpha\right)\right]
\label{eq:mixed_level}
\end{equation}

\noindent where $\alpha$ indicates the neural architecture, $\theta$ is the weight assigned to it, and $\lambda$ is a non-negative regularization variable to control the weights of the training loss and validation loss. When $\lambda=0$, Eq. \ref{eq:mixed_level} reduces to a single-level optimization (Eq. \ref{eq:single_level}); in contrast, Eq. \ref{eq:mixed_level} becomes a bilevel optimization (Eq. \ref{eq:bilevel}). The experimental results presented in \cite{MiLeNAS} showed that mixed-level optimization not only overcomes the overfitting issue of single-level optimization but also avoids the gradient error of bilevel optimization.

Second, in DARTS, the output of each edge is the weighted sum of all candidate operations (shown in Eq. \ref{eq:darts_softmax}) during the whole search stage, which leads to a linear increase in the requirements of GPU memory with the number of candidate operations. To reduce resource consumption, many subsequent studies \cite{gdas,snas,MiLeNAS,wu2018mixed,fbnet} have developed a differentiable sampler to sample a child architecture from the supernet by using a reparameterization trick, namely Gumbel Softmax \cite{gumbel_softmax}. The neural architecture is fully factorized and modeled with a concrete distribution \cite{concrete_dist}, which provides an efficient approach to sampling a child architecture and allows gradient backpropagation. Therefore, Eq. \ref{eq:darts_softmax} is re-formulated as


\begin{equation}
    \overline{o}^k_{i,j}(x)=\sum_{k=1}^K\frac{\exp \left(\left(\log {\alpha}_{i, j}^{k}+{G}_{i, j}^{k}\right) / \tau\right)}{\sum_{l=1}^{K} \exp \left(\left(\log {\alpha}_{i, j}^{l}+{G}_{i, j}^{l}\right) / \tau\right)}o^k(x)
    \label{eq:gumbel_softmax}
\end{equation}

\noindent where ${G}_{i, j}^{k}=-log(-log(u_{i,j}^k))$ is the $k$-th Gumbel sample, $u_{i,j}^k$ is a uniform random variable, and $\tau$ is the Softmax temperature. When $\tau \rightarrow \infty$, the possibility distribution of all operations between each node pair approximates to one-hot distribution. In GDAS \cite{gdas}, only the operation with the maximum possibility for each edge is selected during the forward pass, while the gradient is backpropagated according to Eq. \ref{eq:gumbel_softmax}. In other words, only one path of the supernet is selected for training, thereby reducing the GPU memory usage. Besides, ProxylessNAS \cite{proxylessnas} alleviates the huge resource consumption through path binarization. Specifically, it transforms the real-valued path weights \cite{darts_liu18} to binary gates, which activates only one path of the mixed operations, and hence, solves the memory issue. 


Another problem is the optimization of different operations together, as they may compete with each other, leading to a negative influence. For example, several studies \cite{darts+,pdarts} have found that skip-connect operation dominates at a later search stage in DARTS, which causes the network to be shallower and leads to a marked deterioration in performance. To solve this problem, DARTS+ \cite{darts+} uses an additional early-stop criterion, such that when two or more skip-connects occur in a normal cell, the search process stops. In another example, P-DARTS \cite{pdarts} regularizes the search space by executing operation-level dropout to control the proportion of skip-connect operations occurring during training and evaluation.

\subsubsection{Surrogate Model-based Optimization}


Another group of architecture optimization methods is surrogate model-based optimization (SMBO) algorithms \cite{towards_autotune,towards_nas_hpo,nasbot,deeparchitect,mpas,bayenn,bananas,bayesian_incremental,epnas,pnas_liu18,deeparchitect}. The core concept of SMBO is that it builds a surrogate model of the objective function by iteratively keeping a record of past evaluation results, and uses the surrogate model to predict the most promising architecture. Thus, these methods can substantially shorten the search time and improve efficiency.


SMBO algorithms differ from the surrogate models, which can be broadly divided into Bayesian optimization (BO) methods (including \textit{Gaussian process (GP)} \cite{GP}, \textit{random forest (RF)} \cite{smbo_random_forest}, \textit{tree-structured Parzen estimator (TPE)} \cite{alg_hpo}), and neural networks \cite{bananas,nao,pnas_liu18,epnas}. 

BO \cite{bo,bo_review} is one of the most popular methods for hyperparameter optimization. Many recent studies \cite{towards_autotune,towards_nas_hpo,nasbot,deeparchitect,mpas,bayenn,bananas,bayesian_incremental} have attempted to apply these SOTA BO methods to AO. For example, in \cite{scalable_bo,snoek2012practical,nasbot,bayesian_incremental,smbo_gp_kernel,raiders_gp}, the validation results of the generated neural architectures were modeled as a Gaussian process, which guides the search for the optimal neural architectures. However, in GP-based BO methods, the inference time scales cubically in the number of observations, and they cannot effectively handle variable-length neural networks. Camero et al. \cite{bayesian_free} proposed three fixed-length encoding schemes to cope with variable-length problems by using RF as the surrogate model. Similarly, both \cite{towards_autotune} and \cite{bayesian_free} used RF as a surrogate model, and \cite{auto_weak} showed that it works better in setting high dimensionality than GP-based methods. 


Instead of using BO, some studies have used a neural network as the surrogate model. For example, in PNAS \cite{pnas_liu18} and EPNAS \cite{epnas}, an LSTM is derived as the surrogate model to progressively predict variable-sized architectures. Meanwhile, NAO \cite{nao} uses a simpler surrogate model, i.e., multilayer perceptron (MLP), and NAO is more efficient and achieves better results on CIFAR-10 than does PNAS \cite{pnas_liu18}. White et al. \cite{bananas} trained an ensemble of neural networks to predict the mean and variance of the validation results for candidate neural architectures.

\subsubsection{Grid and Random Search}

Both grid search (GS) and random search (RS) are simple optimization methods applied to several NAS studies \cite{sharpdarts,active_grid_nas,nas_random,eval_nas_phase}. For instance, Geifman et al. \cite{active_grid_nas} proposed a modular architecture search space ($\mathcal{A}=\{A(B,i,j)|i\in\{1,2,...,N_{cells}\}, j\in\{1,2,...,N_{blocks}\} \}$) that is spanned by the grid defined by the two corners $A(B,1,1)$ and $A(B,N_{cells},N_{blocks})$, where $B$ is a searched block structure. Evidently, a larger value $N_{cells}\times N_{blocks}$ leads to the exploration of a larger space, but requires more resources.

The authors in \cite{nas_random} conducted an effectiveness comparison between SOTA NAS methods and RS. The results showed that RS is a competitive NAS baseline. Specifically, RS with an early-stopping strategy performs as well as ENAS \cite{enas}, which is an RL-based leading NAS method. Besides, Yu et al. \cite{eval_nas_phase} demonstrated that the SOTA NAS techniques are not significantly better than random search.

\subsubsection{Hybrid Optimization Method}

The abovementioned architecture optimization methods have their own advantages and disadvantages. 1) EA is a mature global optimization method with high robustness. However, it requires considerable computational resources \cite{amoebanet,large_scale_evolve}, and its evolution operations (such as crossover and mutations) are performed randomly. 2) Although RL-based methods (e.g., ENAS \cite{enas}) can learn complex architectural patterns, the searching efficiency and stability of the RL controller are not guaranteed because it may take several actions to obtain a positive reward. 3) The GD-based methods (e.g., DARTS \cite{darts_liu18}) substantially improve the searching efficiency by relaxing the categorical candidate operations to continuous variables. Nevertheless, in essence, they all search for a child network from a supernet, which limits the diversity of neural architectures. Therefore, some methods have been proposed to incorporate different optimization methods to capture the best of their advantages; these methods are summarized as follows

\textbf{EA+RL}. Chen et al. \cite{reinf_evolv} integrated reinforced mutations into an EA, which avoids the randomness of evolution and improves the searching efficiency. Another similar method developed in parallel is the evolutionary-neural hybrid controller (Evo-NAS) \cite{evolv_hyrid_agent}, which also captures the merits of both RL-based methods and EA. The Evo-NAS controller's mutations are guided by an RL-trained neural network, which can explore a vast search space and sample architectures efficiently. 

\textbf{EA+GD}. Yang et al. \cite{cars} combined the EA and GD-based method. The architectures share parameters within one supernet and are tuned on the training set with a few epochs. Then, the populations and the supernet are directly inherited in the next generation, which substantially accelerates the evolution.
The authors in \cite{cars} only took 0.4 GPU days for searching, which is more efficient than early EA methods (e.g., AmoebaNet \cite{amoebanet} took 3150 GPU days and 450 GPUs for searching). 

\textbf{EA+SMBO}. The authors in \cite{surrogate_evolv} used RF as a surrogate to predict model performance, which accelerates the fitness evaluation in EA.

\textbf{GD+SMBO}. Unlike DARTS, which learns weights for candidate operations, NAO \cite{nao} proposes a variational autoencoder to generate neural architectures and further build a regression model as a surrogate to predict the performance of the generated architecture. The encoder maps the representations of the neural architecture to continuous space, and then a predictor network takes the continuous representations of the neural architecture as input and predicts the corresponding accuracy. Finally, the decoder is used to derive the final architecture from a continuous network representation.

\subsection{Hyperparameter Optimization}

Most NAS methods use the same set of hyperparameters for all candidate architectures during the whole search stage; thus, after finding the most promising neural architecture, it is necessary to redesign a hyperparameter set and use it to retrain or fine-tune the architecture. As some HPO methods (such as BO and RS) have also been applied in NAS, we will only briefly introduce these methods here.




\subsubsection{Grid and Random Search}

\begin{figure}
    \centering
    \includegraphics[width=0.48\textwidth]{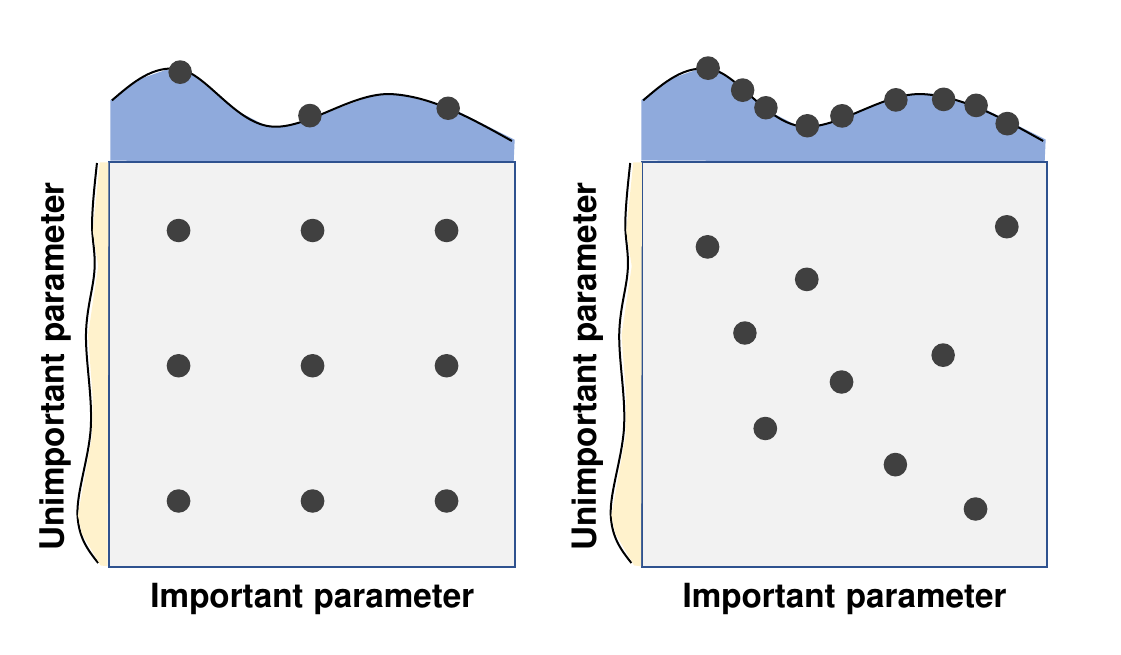}
    \caption{Examples of grid search (left) and random search (right) in nine trials for optimizing a two-dimensional space function $f(x,y)=g(x)+h(y)\approx g(x)$ \cite{hpo_random}. The parameter in $g(x)$ (light-blue part) is relatively important, while that in $h(y)$ (light-yellow part) is not important. In a grid search, nine trials cover only three important parameter values; however, random search can explore nine distinct values of $g$. Therefore, random search is more likely to find the optimal combination of parameters than grid search (the figure is adopted from \cite{hpo_random}).}
    \label{fig:grid_random}
\end{figure}

Figure \ref{fig:grid_random} shows the difference between grid search (GS) and random search (RS): GS divides the search space into regular intervals and selects the best-performing point after evaluating all points; while RS selects the best point from a set of randomly drawn points. 

GS is very simple and naturally supports parallel implementation; however, it is computationally expensive and inefficient when the hyperparameter space is very large, as the number of trials grows exponentially with the dimensionality of hyperparameters. To alleviate this problem, Hsu et al. \cite{hsu2003practical} proposed a coarse-to-fine grid search, in which a coarse grid is first inspected to locate a good region, and then a finer grid search is implemented on the identified region. Similarly, Hesterman et al. \cite{contracting_grid} proposed a contracting GS algorithm, which first computes the likelihood of each point in the grid, and then generates a new grid centered on the maximum-likelihood value. The point separation in the new grid is reduced to half that on the old grid. The above procedure is iterated until the results converge to a local minimum.

Although the authors in \cite{hpo_random} empirically and theoretically showed that RS is more practical and efficient than GS, RS does not promise an optimum value. This means that although a longer search increases the probability of finding optimal hyperparameters, it consumes more resources. Li and Jamieson et al. \cite{li_hyperband_2016} proposed a \textit{hyperband} algorithm to create a tradeoff between the performance of the hyperparameters and resource budgets. The hyperband algorithm allocates limited resources (such as time or CPUs) to only the most promising hyperparameters, by successively discarding the worst half of the configuration settings long before the training process is finished.

\subsubsection{Bayesian Optimization}

Bayesian optimization (BO) is an efficient method for the global optimization of expensive blackbox functions. In this section, we briefly introduce BO. For an in-depth discussion on BO, we recommend readers to refer to the excellent surveys conducted in \cite{bo_review,bo,hpo_chapter,hpo_survey}.

BO is an SMBO method that builds a probabilistic model mapping from the hyperparameters to the objective metrics evaluated on the validation set. It well balances exploration (evaluating as many hyperparameter sets as possible) and exploitation (allocating more resources to promising hyperparameters). 

\begin{algorithm}
\caption{Sequential Model-Based Optimization}
\label{alg:SMBO}
\begin{algorithmic}
\STATE $\text{INPUT:} \, f, \Theta, S, \mathcal{M}$
\STATE ${\mathcal{D} \leftarrow \text { INITSAMPLES }(f, \Theta)}$
\FOR {$i$ in $[1,2,..,T]$}
\STATE ${p(y | \theta, \mathcal{D}) \leftarrow \text { FITMODEL }(\mathcal{M}, \mathcal{D})}$
\STATE ${\theta_{i}\leftarrow\arg\max _{\theta\in \Theta} S(\theta, p(y|\theta, \mathcal{D}))}$
\STATE ${y_{i} \leftarrow f\left(\theta_{i}\right)} \,\,  \,\,  \,\, \triangleright \text{Expensive step}$
\STATE $\mathcal{D} \leftarrow \mathcal{D} \cup\left(\theta_{i}, y_{i}\right)$
\ENDFOR
\end{algorithmic}
\end{algorithm}

The steps of SMBO are expressed in Algorithm \ref{alg:SMBO} (adopted from \cite{bo}). Here, several inputs need to be predefined initially, including an evaluation function $f$, search space $\Theta$, acquisition function $\mathcal{S}$, probabilistic model $\mathcal{M}$, and record dataset $D$. Specifically, $D$ is a dataset that records many sample pairs ($\theta_i, y_i$), where $\theta_i\in \Theta$ indicates a sampled neural architecture and $y_i$ indicates its evaluation result. After the initialization, the SMBO steps are described as follows:

\begin{enumerate}
    \item The first step is to tune the probabilistic model $\mathcal{M}$ to fit the record dataset $D$.
    \item The acquisition function $\mathcal{S}$ is used to select the next promising neural architecture from the probabilistic model $\mathcal{M}$.
    \item The performance of the selected neural architecture is evaluated by $f$, which is an expensive step as it involves training the neural network on the training set and evaluating it on the validation set.
    \item The record dataset $D$ is updated by appending a new pair of results $(\theta_i,y_i)$.
\end{enumerate}


The above four steps are repeated $T$ times, where $T$ needs to be specified according to the total time or resources available. The commonly used surrogate models for the BO method are GP, RF, and TPE. Table \ref{table:bo_models} summarizes the existing open-source BO methods, where GP is one of the most popular surrogate models. However, GP scales cubically with the number of data samples, while RF can natively handle large spaces and scales better to many data samples. Besides, Falkner and Klein et al. \cite{bohb} proposed the BO-based hyperband (BOHB) algorithm, which combines the strengths of TPE-based BO and hyperband, and hence, performs much better than standard BO methods. Furthermore, FABOLAS \cite{fabolas} is a faster BO procedure, which maps the validation loss and training time as functions of dataset size, i.e., trains a generative model on a sub-dataset that gradually increases in size. Here, FABOLAS is 10-100 times faster than other SOTA BO algorithms and identifies the most promising hyperparameters.

\begin{table}[]
    \centering
    \begin{tabular}{c|c}
    \hline
    \textbf{Library}  & \textbf{Model}      \\ \hline
    \makecell{ Spearmint\\ https://github.com/HIPS/Spearmint}   & GP \\\hline
    \makecell{ MOE\\ https://github.com/Yelp/MOE}   &  GP\\\hline
    \makecell{ PyBO\\ https://github.com/mwhoffman/pybo}   & GP \\\hline
    \makecell{ Bayesopt\\ https://github.com/rmcantin/bayesopt}   &GP  \\\hline
    \makecell{ SkGP\\ https://scikit-optimize.github.io}   & GP \\\hline
    \makecell{ GPyOpt\\ http://sheffieldml.github.io/GPyOpt}   & GP \\\hline
    
    \makecell{SMAC      \\ https://github.com/automl/SMAC3}   & RF    \\\hline
    \makecell{Hyperopt  \\ http://hyperopt.github.io/hyperopt} & TPE  \\ \hline
    \makecell{BOHB\\https://github.com/automl/HpBandSter}&TPE \\\hline
\end{tabular}
    \caption{Open-source Bayesian optimization libraries. GP, RF, and TPE represent \textit{Gaussian process} \cite{GP}, \textit{random forest} \cite{smbo_random_forest}, and \textit{tree-structured Parzen estimator} \cite{alg_hpo}, respectively. }
    \label{table:bo_models}
\end{table}

\subsubsection{Gradient-based Optimization}


Another group of HPO methods are \textit{gradient-based optimization (GO)} algorithms \cite{go_hpo,generic_opt,gd_hpo_revers,hpo_appro_gd,forward_gd_hpo,gd_ultimate_hpo}. Unlike the above blackbox HPO methods (e.g., GS, RS, and BO), GO methods use the gradient information to optimize the hyperparameters and substantially improve the efficiency of HPO. Maclaurin et al. \cite{gd_hpo_revers} proposed a reversible-dynamics memory-tape approach to handle thousands of hyperparameters efficiently through the gradient information. However, optimizing many hyperparameters is computationally challenging. To alleviate this issue, the authors in \cite{hpo_appro_gd} used approximate gradient information rather than the true gradient to optimize continuous hyperparameters, where the hyperparameters can be updated before the model is trained to converge. Franceschi et al. \cite{forward_gd_hpo} studied both reverse- and forward-mode GO methods. The reverse-mode method differs from the method proposed in \cite{gd_hpo_revers} and does not require reversible dynamics; however, it needs to store the entire training history for computing the gradient with respect to the hyperparameters. The forward-mode method overcomes this problem by using real-time updating hyperparameters, and is demonstrated to significantly improve the efficiency of HPO on large datasets. Chandra \cite{gd_ultimate_hpo} proposed a gradient-based ultimate optimizer, which can optimize not only the regular hyperparameters (e.g., learning rate) but also those of the optimizer (e.g., Adam optimizer \cite{kingma2014adam}'s moment coefficient $\beta_1,\beta_2$).


% ...

\section{Model Evaluation}\label{section:model_estimation}

Once a new neural network has been generated, its performance must be evaluated. An intuitive method is to train the network to convergence and then evaluate its performance. However, this method requires extensive time and computing resources. For example, \cite{nas_rl_zoph16} took 800 K40 GPUs and 28 days in total to search. Additionally, NASNet \cite{nasnet_zoph17} and AmoebaNet \cite{amoebanet} required 500 P100 GPUs and 450 K40 GPUs, respectively. In this section, we summarize several algorithms for accelerating the process of model evaluation.


\subsection{Low fidelity}

As model training time is highly related to the dataset and model size, model evaluation can be accelerated in different ways. First, the number of images or the resolution of images (in terms of image-classification tasks) can be decreased. For example, FABOLAS \cite{fabolas} trains the model on a subset of the training set to accelerate model evaluation. In \cite{imagenet_variants}, ImageNet64$\times$64 and its variants 32$\times$32, 16$\times$16 are provided, while these lower resolution datasets can retain characteristics similar to those of the original ImageNet dataset. Second, low-fidelity model evaluation can be realized by reducing the model size, such as by training with fewer filters per layer \cite{nasnet_zoph17, amoebanet}. By analogy to ensemble learning, \cite{multi_fidelity} proposes the \textit{Transfer Series Expansion (TSE)}, which constructs an ensemble estimator by linearly combining a series of basic low-fidelity estimators, hence avoiding the bias that can derive from using a single low-fidelity estimator. Furthermore, Zela et al. \cite{towards_nas_hpo} empirically demonstrated that there is a weak correlation between performance after short or long training times, thus confirming that a prolonged search for network configurations is unnecessary.

\subsection{Weight sharing} 

In \cite{nas_rl_zoph16}, once a network has been evaluated, it is dropped. Hence, the technique of weight sharing is used to accelerate the process of NAS. For example, Wong and Lu et al. \cite{wong2018transfer} proposed transfer neural AutoML, which uses knowledge from prior tasks to accelerate network design. ENAS \cite{enas} shares parameters among child networks, leading to a thousand-fold faster network design than \cite{nas_rl_zoph16}. Network morphism based algorithms \cite{chen2015net2net,network_morphism_wei16} can also inherit the weights of previous architectures, and single-path NAS \cite{singlepath_nas} uses a single-path over-parameterized ConvNet to encode all architectural decisions with shared convolutional kernel parameters.

\subsection{Surrogate}

The surrogate-based method \cite{eggensperger2014surrogate,wang2014evaluation,eggensperger2015efficient,surrogate_evolv} is another powerful tool that approximates the black-box function. In general, once a good approximation has been obtained, it is trivial to find the configurations that directly optimize the original expensive objective. For example, Progressive Neural Architecture Search (PNAS) \cite{pnas_liu18} introduces a surrogate model to control the method of searching. Although ENAS has been proven to be very efficient, PNAS is even more efficient, as the number of models evaluated by PNAS is over five times that evaluated by ENAS, and PNAS is eight times faster in terms of total computational speed. A well-performing surrogate usually requires large amounts of labeled architectures, while the optimization space is too large and hard to quantify, and the evaluation of each configuration is extremely expensive \cite{smbo_vu17}. To alleviate this issue, Luo et al. \cite{semi_NAS} proposed \textit{SemiNAS}, a semi-supervised NAS method, which leverages amounts of unlabeled architectures to train the surrogate, a controller that is used to predict the accuracy of architectures without evaluation. Initially, the surrogate is only trained with a small number of labeled data pairs \textit{(architectures, accuracy)}, then the generated data pairs will be gradually added to the original data to further improve the surrogate.

\subsection{Early stopping}

Early stopping was first used to prevent overfitting in classical ML, and it has been used in several recent studies \cite{learn_curve_Klein16,peephole,speedup_learn_curve} to accelerate model evaluation by stopping evaluations that are predicted to perform poorly on the validation set. For example, \cite{speedup_learn_curve} proposes a learning-curve model that is a weighted combination of a set of parametric curve models selected from the literature, thereby enabling the performance of the network to be predicted. Furthermore, \cite{earlystop_no_val} presents a novel approach for early stopping based on fast-to-compute local statistics of the computed gradients, which no longer relies on the validation set and allows the optimizer to make full use of all of the training data.

\section{NAS Discussion}\label{section:nas_performance}


\begin{table*}[]
\centering
\scalebox{0.94}{
\begin{tabular}{l|c|c|c|c|c|c}
\hline
\multicolumn{1}{c|}{\textbf{Reference}} &  \tabincell{c}{\textbf{Published}\\\textbf{in}}&
\tabincell{c}{\textbf{\#Params} \\ \textbf{(Millions)}} &  \tabincell{c}{\textbf{Top-1} \\\textbf{Acc(\%)}}  &    \tabincell{c}{\textbf{GPU} \\ \textbf{Days}}  & \tabincell{c}{\textbf{\#GPUs}}& \tabincell{c}{\textbf{AO}}\\ 
\hline

ResNet-110 \cite{resnet} & ECCV16&1.7 &93.57 & -&-& \\
PyramidNet \cite{pyramidnet} & CVPR17& 26& 96.69& -&-& \\
DenseNet \cite{densenet} & CVPR17& 25.6&96.54 & -&-& \multirow{-4}{*}{\makecell{Manually\\designed}}\\
\hline

GeNet\#2 (G-50)  \cite{genertic_cnn} & ICCV17 & - & 92.9    & 17 & - &  \\
Large-scale ensemble  \cite{large_scale_evolve} & ICML17 & 40.4 & 95.6  & 2,500 &250 &  \\
Hierarchical-EAS \cite{liu_hierarchical_2018}  & ICLR18 & 15.7 & 96.25 & 300    & 200  & \\
CGP-ResSet  \cite{cgp_suganuma17}  & IJCAI18  & 6.4    & 94.02   & 27.4& 2 &  \\
AmoebaNet-B (N=6, F=128)+c/o   \cite{amoebanet}  &  AAAI19  & 34.9 & 97.87 & 3,150 & 450 K40&  \\
AmoebaNet-B (N=6, F=36)+c/o  \cite{amoebanet}  &  AAAI19  & 2.8  & 97.45 & 3,150& 450 K40&  \\
Lemonade  \cite{lemonade} & ICLR19  & 3.4 & 97.6 & 56  & 8 Titan & \\
EENA \cite{eena}  & ICCV19 & 8.47 & 97.44 & 0.65 & 1 Titan Xp & \\
EENA (more channels)\cite{eena} & ICCV19  & 54.14 & 97.79 & 0.65 & 1 Titan Xp & 
\multirow{-9}{*}{\makecell{EA}} \\ 
\hline

NASv3\cite{nas_rl_zoph16}  & ICLR17 & 7.1    & 95.53   & 22,400  & 800 K40    &  \\
NASv3+more filters \cite{nas_rl_zoph16} & ICLR17 & 37.4 & 96.35 &22,400 &800 K40 & \\
MetaQNN   \cite{metaqnn}  & ICLR17 & - & 93.08   & 100& 10&  \\
NASNet-A (7 @ 2304)+c/o  \cite{nasnet_zoph17} & CVPR18 & 87.6 & 97.60 & 2,000 & 500 P100&  \\
NASNet-A (6 @ 768)+c/o \cite{nasnet_zoph17}  & CVPR18  & 3.3  & 97.35 & 2,000 & 500 P100 &  \\
Block-QNN-Connection more filter  \cite{blockqnn} & CVPR18 & 33.3 & 97.65 & 96 &32 1080Ti&  \\
Block-QNN-Depthwise, N=3    \cite{blockqnn} & CVPR18 & 3.3  & 97.42 & 96  & 32 1080Ti&  \\
ENAS+macro  \cite{enas} & ICML18 & 38.0   & 96.13   & 0.32& 1 &  \\
ENAS+micro+c/o   \cite{enas}  & ICML18  & 4.6    & 97.11   & 0.45& 1 & \\
Path-level EAS \cite{path_level_nm} & ICML18  & 5.7 & 97.01 & 200 & - & \\
Path-level EAS+c/o \cite{path_level_nm} & ICML18  & 5.7 & 97.51 & 200 & - & \\
ProxylessNAS-RL+c/o\cite{proxylessnas} & ICLR19  & 5.8 & 97.70 & - & - & \\
FPNAS\cite{fpnas}&ICCV19&5.76&96.99&-&-&
\multirow{-13}{*}{RL} \\ 
\hline

DARTS(first order)+c/o\cite{darts_liu18} &  ICLR19 & 3.3 & 97.00   & 1.5  & 4 1080Ti&  \\
DARTS(second order)+c/o\cite{darts_liu18}  & ICLR19 & 3.3 & 97.23   & 4  & 4 1080Ti&  \\
sharpDARTS   \cite{sharpdarts} & ArXiv19 & 3.6    & 98.07   & 0.8& 1 2080Ti&  \\
P-DARTS+c/o\cite{pdarts} & ICCV19 & 3.4 & 97.50 & 0.3 & -   &  \\
P-DARTS(large)+c/o\cite{pdarts} & ICCV19 &10.5&97.75 &0.3 & - &  \\
SETN\cite{oneshot_self}&ICCV19&4.6&97.31&1.8&-&\\
GDAS+c/o \cite{gdas}  & CVPR19 & 2.5 & 97.18 & 0.17 &  1 & \\
SNAS+moderate constraint+c/o  \cite{snas} & ICLR19 & 2.8  & 97.15 & 1.5& 1 & \\
BayesNAS\cite{bayesnas}  & ICML19 &3.4&97.59&0.1&1&\\
ProxylessNAS-GD+c/o\cite{proxylessnas} &  ICLR19 & 5.7 & 97.92 & - & -& \\
PC-DARTS+c/o \cite{xu2019pcdarts}  & CVPR20 & 3.6 & 97.43 & 0.1 & 1 1080Ti & \\
MiLeNAS\cite{MiLeNAS} & CVPR20 &3.87&97.66&0.3&-&\\
SGAS\cite{sgas} & CVPR20 &3.8&97.61&0.25&1 1080Ti&\\
GDAS-NSAS\cite{nsas}&CVPR20&3.54&97.27&0.4&-&\multirow{-16}{*}{\makecell{GD}}  \\ 
\hline

NASBOT\cite{nasbot}&NeurIPS18&-&91.31&1.7&-&\\
PNAS \cite{pnas_liu18}  &  ECCV18 & 3.2   & 96.59   & 225  & -&\\
EPNAS\cite{epnas} & BMVC18 &6.6&96.29&1.8&1&\\
GHN\cite{GHN}&ICLR19&5.7&97.16&0.84&-&
\multirow{-4}{*}{\makecell{SMBO}}\\ 
\hline

NAO+random+c/o\cite{nao} & NeurIPS18 &10.6 & 97.52 & 200&200 V100&\\
SMASH \cite{smash_onehot}  & ICLR18 & 16 & 95.97 & 1.5 & - & \\
Hierarchical-random  \cite{liu_hierarchical_2018} & ICLR18 & 15.7   & 96.09   & 8  & 200    &  \\
RandomNAS    \cite{nas_random} & UAI19 & 4.3  & 97.15 & 2.7    & -    &  \\
DARTS - random+c/o    \cite{darts_liu18}  & ICLR19  & 3.2    & 96.71   & 4  & 1 & \\
RandomNAS-NSAS\cite{nsas}&CVPR20&3.08&97.36&0.7&-&
\multirow{-6}{*}{\makecell{RS}}\\ 
\hline

NAO+weight sharing+c/o  \cite{nao}  & NeurIPS18  & 2.5  & 97.07 & 0.3 & 1 V100  &  GD+SMBO\\
RENASNet+c/o\cite{reinf_evolv} & CVPR19 &3.5&91.12&1.5&4&EA+RL\\
CARS\cite{cars} & CVPR20 &3.6&97.38&0.4&-& EA+GD
\multirow{-2}{*}{\makecell{}}\\ \hline

\end{tabular}
}
\caption{
Performance of different NAS algorithms on CIFAR-10. The ``AO'' column indicates the architecture optimization method. The \textit{dash (-)} indicates that the corresponding information is not provided in the original paper. ``c/o'' indicates the use of Cutout \cite{cutout}. RL, EA, GD, RS, and SMBO indicate reinforcement learning, evolution-based algorithm, gradient descent, random search, and surrogate model-based optimization, respectively.}
\label{table:NAS_cifar_perf}
\end{table*}

\begin{table*}[!t]
\centering
\begin{tabular}{l|c|c|c|c|c|c}
\hline
\multicolumn{1}{c|}{\textbf{Reference}} &  \tabincell{c}{\makecell{\textbf{Published}\\\textbf{in}}}&
\tabincell{c}{\textbf{\#Params} \\ \textbf{(Millions)}} &  \tabincell{c}{\textbf{Top-1/5} \\\textbf{Acc(\%)}}  &    \tabincell{c}{\textbf{GPU} \\ \textbf{Days}}  & \tabincell{c}{\textbf{\#GPUs}}& \tabincell{c}{\textbf{AO}}    \\  \hline

ResNet-152 \cite{resnet} & CVPR16& 230 &70.62/95.51 & -&-& \\
PyramidNet \cite{pyramidnet} & CVPR17&116.4 & 70.8/95.3& -&-& \\
SENet-154 \cite{senet} & CVPR17& -&71.32/95.53 & -&-& \\
DenseNet-201 \cite{densenet} & CVPR17& 76.35&78.54/94.46 & -&-&\\
MobileNetV2 \cite{MobileNetV2} &CVPR18&6.9&74.7/-&-&-& \multirow{-4}{*}{\makecell{Manually\\designed}}\\
\hline

GeNet\#2\cite{genertic_cnn} & ICCV17&-&72.13/90.26&17&-&\\
AmoebaNet-C(N=4,F=50)\cite{amoebanet} & AAAI19&6.4&75.7/92.4&3,150&450 K40&\\
Hierarchical-EAS\cite{liu_hierarchical_2018} & ICLR18&-&79.7/94.8&300&200&\\
AmoebaNet-C(N=6,F=228)\cite{amoebanet} & AAAI19&155.3&83.1/96.3&3,150&450 K40&\\
GreedyNAS \cite{you2020greedynas}&CVPR20&6.5&77.1/93.3&1&-&
\multirow{-4}{*}{EA} \\ \hline

NASNet-A(4@1056) & ICLR17 & 5.3 & 74.0/91.6 & 2,000&500 P100&\\
NASNet-A(6@4032) & ICLR17 & 88.9 & 82.7/96.2 & 2,000&500 P100&\\
Block-QNN\cite{blockqnn}  & CVPR18&91& 81.0/95.42 &96&32 1080Ti&\\
Path-level EAS\cite{path_level_nm}  &ICML18 & - & 74.6/91.9 & 8.3 & - & \\
ProxylessNAS(GPU) \cite{proxylessnas}  &ICLR19 & - & 75.1/92.5 & 8.3 & - & \\
ProxylessNAS-RL(mobile) \cite{proxylessnas}  &ICLR19 & - & 74.6/92.2 & 8.3 & - & \\
MnasNet\cite{mnasnet}& CVPR19& 5.2&76.7/93.3&1,666&-&\\
EfficientNet-B0\cite{efficientnet} & ICML19&5.3&77.3/93.5&-&-&\\
EfficientNet-B7\cite{efficientnet} & ICML19&66&84.4/97.1&-&-&\\
FPNAS\cite{fpnas}&ICCV19&3.41&73.3/-&0.8&-&
\multirow{-10}{*}{RL} \\ \hline

DARTS (searched on CIFAR-10)\cite{darts_liu18}  & ICLR19& 4.7 & 73.3/81.3 & 4 & - & \\
sharpDARTS\cite{sharpdarts} & Arxiv19&4.9&74.9/92.2&0.8&-&\\
P-DARTS\cite{pdarts} & ICCV19&4.9&75.6/92.6&0.3&-&\\
SETN\cite{oneshot_self}&ICCV19&5.4&74.3/92.0&1.8&-&\\
GDAS \cite{gdas}  &CVPR19 & 4.4 & 72.5/90.9 & 0.17 & 1 &\\
SNAS\cite{snas}&ICLR19 & 4.3&72.7/90.8&1.5&-&\\
ProxylessNAS-G\cite{proxylessnas}&ICLR19&-&74.2/91.7&-&-&\\
BayesNAS\cite{bayesnas}  & ICML19&3.9&73.5/91.1&0.2&1&\\
FBNet\cite{fbnet} &CVPR19 &5.5&74.9/-&216&-&\\
OFA\cite{OFA}&ICLR20&7.7&77.3/-&-&-&\\
AtomNAS\cite{atomnas}&ICLR20&5.9&77.6/93.6&-&-&\\
MiLeNAS\cite{MiLeNAS}& CVPR20& 4.9&75.3/92.4&0.3&-&\\
DSNAS\cite{hu2020dsnas}&CVPR20&-&74.4/91.54&17.5&4 Titan X&\\
SGAS\cite{sgas}& CVPR20& 5.4&75.9/92.7&0.25&1 1080Ti&\\
PC-DARTS \cite{xu2019pcdarts} & CVPR20 & 5.3 & 75.8/92.7 & 3.8 & 8 V100 & \\
DenseNAS\cite{densenas}&CVPR20&-&75.3/-&2.7&-&\\
FBNetV2-L1\cite{fbnetv2}&CVPR20&-&77.2/-&25&8 V100&
\multirow{-15}{*}{GD} \\ \hline

PNAS-5(N=3,F=54)\cite{pnas_liu18}&ECCV18 & 5.1&74.2/91.9&225&-&\\
PNAS-5(N=4,F=216)\cite{pnas_liu18} & ECCV18&86.1&82.9/96.2&225&-&\\
GHN\cite{GHN}&ICLR19&6.1&73.0/91.3&0.84&-&\\
SemiNAS\cite{semi_NAS}&CVPR20&6.32&76.5/93.2&4&-&
\multirow{-4}{*}{SMBO}\\ \hline

Hierarchical-random\cite{liu_hierarchical_2018}& ICLR18& -&79.6/94.7&8.3&200&\\
OFA-random\cite{OFA}&CVPR20&7.7&73.8/-&-&-&
\multirow{-2}{*}{RS}\\ \hline

RENASNet\cite{reinf_evolv}& CVPR19& 5.36&75.7/92.6&-&-&EA+RL\\
Evo-NAS\cite{evolv_hyrid_agent}& Arxiv20& -&75.43/-&740&-&EA+RL\\
CARS\cite{cars} & CVPR20&5.1&75.2/92.5&0.4&-&EA+GD\multirow{-2}{*}{}\\ \hline

\end{tabular}
\caption{
Performance of different NAS algorithms on ImageNet. The ``AO'' column indicates the architecture optimization method. The \textit{dash (-)} indicates that the corresponding information is not provided in the original paper. RL, EA, GD, RS, and SMBO indicate reinforcement learning, evolution-based algorithm, gradient descent, random search, and surrogate model-based optimization, respectively.}
\label{table:NAS_imagenet_perf}
\end{table*}

In Section \ref{section:model_generation}, we reviewed the various search space and architecture optimization methods, and in Section \ref{section:model_estimation}, we summarized commonly used model evaluation methods. These two sections introduced many NAS studies, which may cause the readers to get lost in details. Therefore, in this section, we summarize and compare these NAS algorithms' performance from a global perspective to provide readers a clearer and more comprehensive understanding of NAS methods' development. Then, we discuss some major topics of the NAS technique.

\subsection{NAS Performance Comparison}

Many NAS studies have proposed several neural architecture variants, where each variant is designed for different scenarios. For instance, some architecture variants perform better but are larger, while some are lightweight for a mobile device but with a performance penalty. Therefore, we only report the representative results of each study. Besides, to ensure a valid comparison, we consider the accuracy and algorithm efficiency as comparison indices. As the number and types of GPUs used vary for different studies, we use \textit{GPU Days} to approximate the efficiency, which is defined as:

\begin{equation}
    \text{GPU Days}=N \times D
    \label{eq:gpu_days}
\end{equation}

\noindent where $N$ represents the number of GPUs, and $D$ represents the actual number of days spent searching. 

Tables \ref{table:NAS_cifar_perf} and \ref{table:NAS_imagenet_perf} present the performances of different NAS methods on CIFAR-10 and ImageNet, respectively. Besides, as most NAS methods first search for the neural architecture based on a small dataset (CIFAR-10), and then transfer the architecture to a larger dataset (ImageNet), the search time for both datasets is the same. The tables show that the early studies on EA- and RL-based NAS methods focused more on high performance, regardless of the resource consumption. For example, although AmoebaNet \cite{amoebanet} achieved excellent results for both CIFAR-10 and ImageNet, the searching took 3,150 GPU days and 450 GPUs. The subsequent NAS studies attempted to improve the searching efficiency while ensuring the searched model’s high performance. For instance, EENA \cite{eena} elaborately designs the mutation and crossover operations, which can reuse the learned information to guide the evolution process, and hence, substantially improve the efficiency of EA-based NAS methods. ENAS \cite{enas} is one of the first RL-based NAS methods to adopt the parameter-sharing strategy, which reduces the number of GPU budgets to 1 and shortens the searching time to less than one day. We also observe that gradient descent-based architecture optimization methods can substantially reduce the computational resource consumption for searching, and achieve SOTA results. Several follow-up studies have been conducted to achieve further improvement and optimization in this direction. Interestingly, RS-based methods can also obtain comparable results. The authors in \cite{nas_random} demonstrated that RS with weight-sharing could outperform a series of powerful methods, such as ENAS \cite{enas} and DARTS \cite{darts_liu18}.

\subsubsection{Kendall Tau Metric}

As RS is comparable to more sophisticated methods (e.g., DARTS and ENAS), a natural question is, what are the advantages and significance of the other AO algorithms compared with RS? Researchers have tried to use other metrics to answer this question, rather than simply considering the model's final accuracy. Most NAS methods comprise two stages: 1) search for a best-performing architecture on the training set and 2) expand it to a deeper architecture and estimate it on the validation set. However, there usually exists a large gap between the two stages. In other words, the architecture that achieves the best result in the training set is not necessarily the best one for the validation set. Therefore, instead of merely considering the final accuracy and search time cost, many NAS studies \cite{hu2020dsnas,istrate2018tapas,nsas,eval_nas_phase,nas_eval} have used Kendall Tau ($\tau$) metric \cite{kendall_tau} to evaluate the correlation of the model performance between the search and evaluation stages. The parameter $\tau$ is defined as

\begin{equation}
    \tau=\frac{N_C-N_D}{N_C+N_D}
\end{equation}

\noindent where $N_C$ and $N_D$ indicate the numbers of concordant and discordant pairs. $\tau$ is a number in the range [-1,1] with the following properties:

\begin{itemize}
    \item $\tau=1$: two rankings are identical
    \item $\tau=-1$: two rankings are completely opposite.
    \item $\tau=0$: there is no relationship between two rankings.
\end{itemize}


\begin{figure}
    \centering
    \subfigure[Two-stage NAS comprises the searching stage and evaluation stage. The best-performing model of the searching stage is further retrained in the evaluation stage.]{\includegraphics[width=0.48\textwidth]{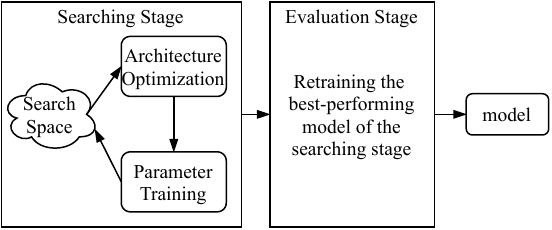}}
    \subfigure[One-stage NAS can directly deploy a well-performing model without extra retraining or fine-tuning. The two-way arrow indicates that the processes of architecture optimization and parameter training run simultaneously. ]{\includegraphics[width=0.48\textwidth]{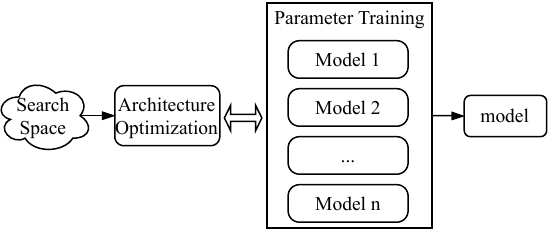}}
    \caption{Illustration of two- and one-stage neural architecture search flow.}
    \label{fig:nas_stage}
\end{figure}

\subsubsection{NAS-Bench Dataset}

Although Tables \ref{table:NAS_cifar_perf} and \ref{table:NAS_imagenet_perf} present a clear comparison between different NAS methods, the results of different methods are obtained under different settings, such as training-related hyperparameters (e.g., batch size and training epochs) and data augmentation (e.g., Cutout \cite{cutout}). In other words, the comparison is not quite fair. In this context, NAS-Bench-101 \cite{ying2019nasbench} is a pioneering work for improving the reproducibility. It provides a tabular dataset containing 423,624 unique neural networks generated and evaluated from a fixed graph-based search space and mapped to their trained and evaluated performance on CIFAR-10. Meanwhile, Dong et al. \cite{nasbench201} further built NAS-Bench-201, which is an extension to NAS-Bench-101 and has a different search space, results on multiple datasets (CIFAR-10, CIFAR-100, and ImageNet-16-120 \cite{imagenet_variants}), and more diagnostic information. Similarly, Klyuchnikov et al. \cite{nasbenchnlp} proposed a NAS-Bench for the NLP task. These datasets enable NAS researchers to focus solely on verifying the effectiveness and efficiency of their AO algorithms, avoiding repetitive training for selected architectures and substantially helping the NAS community to develop.

\subsection{One-stage vs. Two-stage}

The NAS methods can be roughly divided into two classes according to the flow ––two-stage and one-stage–– as shown in Figure \ref{fig:nas_stage}.

\textbf{Two-stage NAS} comprises the \textit{searching stage} and \textit{evaluation stage}. The \textit{searching stage} involves two processes: architecture optimization, which aims to find the optimal architecture, and parameter training, which is to train the found architecture's parameter. The simplest idea is to train all possible architectures' parameters from scratch and then choose the optimal architecture. However, it is resource-consuming (e.g., NAS-RL \cite{nas_rl_zoph16} took 22,400 GPU days with 800 K40 GPUs for searching) ), which is infeasible for most companies and institutes. Therefore, most NAS methods (such as ENAS \cite{enas} and DARTS \cite{darts_liu18}) sample and train many candidate architectures in the searching stage, and then further retrain the best-performing architecture in the evaluation stage.

\textbf{One-stage NAS} refers to a class of NAS methods that can export a well-designed and well-trained neural architecture without extra retraining, by running AO and parameter training simultaneously. In this way, the efficiency can be substantially improved. However, model architecture and its weight parameters are highly coupled; it is difficult to optimize them simultaneously. Several recent studies  \cite{OFA,yoso,yu2020bignas,atomnas} have attempted to overcome this challenge. For instance, the authors in \cite{OFA} proposed the \textit{progressive shrinking} algorithm to post-process the weights after the training was completed. They first pretrained the entire neural network, and then progressively fine-tuned the smaller networks that shared weights with the complete network. Based on well-designed constraints, the performance of all subnetworks was guaranteed. Thus, given a target deployment device, a specialized subnetwork can be directly exported without fine-tuning. However, \cite{OFA} was still computational resource-intensive, as the whole process took 1,200 GPU hours with V100 GPUs. BigNAS \cite{yu2020bignas} revisited the conventional training techniques of stand-alone networks, and empirically proposed several techniques to handle a wider set of models, ranging in size from 200M to 1G FLOPs, whereas \cite{OFA} only handled models under 600M FLOPs. Both AtomNAS \cite{atomnas} and DSNAS \cite{hu2020dsnas} proposed an end-to-end one-stage NAS framework to further boost the performance and simplify the flow.

\subsection{One-shot/Weight-sharing}

\begin{figure}
    \centering
    \includegraphics[width=0.48\textwidth]{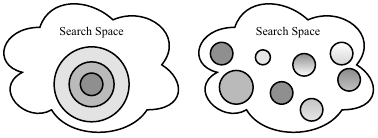}
    \caption{(Left) One-shot models. (Right) Non-one-shot models. Each circle indicates a different model, and its area indicates the model's size. We use concentric circles to represent one-shot models, as they share the weights with each other.}
    \label{fig:one-shot}
\end{figure}

\textbf{One-shot$\neq$one-stage.} Note that one shot is not exactly equivalent to one stage. As mentioned above, we divide the NAS studies into one- and two-stage methods according to the flow (Figure \ref{fig:nas_stage}), whereas whether a NAS algorithm belongs to a one-shot method depends on whether the candidate architectures share the same weights (Figure \ref{fig:one-shot}). However, we observe that most one-stage NAS methods are based on the one-shot paradigm. 

\textbf{ What is One-shot NAS?} One-shot NAS methods embed the search space into an overparameterized supernet, and thus, all possible architectures can be derived from the supernet. Figure \ref{fig:nas_stage} shows the difference between the search spaces of one-shot and non-one-shot NAS. Each circle indicates a different architecture, where the architectures of one-shot NAS methods share the same weights. One-shot NAS methods can be divided into two categories according to how to handle AO and parameter training: coupled and decoupled optimization \cite{chu2019fairnas,you2020greedynas}.

\textbf{Coupled optimization.} The first category of one-shot NAS methods optimizes the architecture and weights in a coupled manner \cite{enas,darts_liu18,gdas,proxylessnas,snas}. For instance, ENAS \cite{enas} uses an LSTM network to discretely sample a new architecture, and then uses a few batches of the training data to optimize the weight of this architecture. After repeating the above steps many times, a collection of architectures and their corresponding performances are recorded. Finally, the best-performing architecture is selected for further retraining. DARTS \cite{darts_liu18} uses a similar weight sharing strategy, but has a continuously parameterized architecture distribution. The supernet contains all candidate operations, each with learnable parameters. The best architecture can be directly derived from the distribution. However, as DARTS \cite{darts_liu18} directly optimizes the supernet weights and the architecture distribution, it suffers from vast GPU memory consumption. Although DARTS-like methods \cite{proxylessnas,gdas, snas} have adopted different approaches to reduce the resource requirements, coupled optimization inevitably introduces a bias in both architecture distribution and supernet weights \cite{singlepath_nas,chu2019fairnas}, as they treat all subnetworks unequally. The rapidly converged architectures can easily obtain more opportunities to be optimized \cite{darts_liu18,darts+}, and are only a small portion of all candidates; therefore, it is challenging to find the best architecture.

Another disadvantage of coupling optimization is that when new architectures are sampled and trained continuously, the weights of previous architectures are negatively impacted, leading to performance degradation. The authors in \cite{multimodel_forget} defined this phenomenon as \textit{multimodel forgetting}. To overcome this problem, Zhang et al. \cite{oneshot_forget} modeled supernet training as a constrained optimization problem of continual learning and proposed \textit{novel search-based architecture selection} (NSAS) loss function. They applied the proposed method to RandomNAS \cite{nas_random} and GDAS \cite{gdas}, where the experimental result demonstrated that the method effectively reduces the multimodel forgetting and boosting the predictive ability of the supernet as an evaluator.

\textbf{Decoupled optimization}. The second category of one-shot NAS methods \cite{oneshot_self,under_oneshot,chu2019fairnas,OFA} decouples the optimization of architecture and weights into two sequential phases: 1) training the supernet and 2) using the trained supernet as a predictive performance estimator of different architectures to select the most promising architecture.

In terms of the supernet training phase, the supernet cannot be directly trained as a regular neural network because its weights are also deeply coupled \cite{singlepath_nas}. Yu et al. \cite{eval_nas_phase} experimentally showed that the weight-sharing strategy degrades the individual architecture's performance and negatively impacts the real performance ranking of the candidate architectures. To reduce the weight coupling, many one-shot NAS methods \cite{singlepath_nas,oneshot_self,smash_onehot,GHN} adopt the random sampling policy, which randomly samples an architecture from the supernet, activating and optimizing only the weights of this architecture. Meanwhile, RandomNAS \cite{nas_random} demonstrates that a random search policy is a competitive baseline method. Although some one-shot approaches \cite{gdas,enas,snas,proxylessnas,fbnet} have adopted the strategy that samples and trains only one path of the supernet at a time, they sample the path according to the RL controller \cite{enas}, Gumbel Softmax \cite{gdas,snas,fbnet}, or the BinaryConnect network \cite{proxylessnas}, which instead highly couples the architecture and supernet weights. SMASH \cite{smash_onehot} adopts an auxiliary hypernetwork to generate weights for randomly sampled architectures.
Similarly, Zhang et al. \cite{GHN} proposed a computation graph representation, and used the graph hypernetwork (GHN) to predict the weights for all possible architectures faster and more accurately than regular hypernetworks \cite{smash_onehot}. However, through a careful experimental analysis conducted to understand the weight-sharing strategy's mechanism, Bender et al. \cite{under_oneshot} showed that neither a hypernetwork nor an RL controller is required to find the optimal architecture. They proposed a \textit{path dropout} strategy to alleviate the problem of weight coupling. During supernet training, each path of the supernet is randomly dropped with gradually increasing probability. GreedyNAS \cite{you2020greedynas} adopts a multipath sampling strategy to train the greedy supernet. This strategy focuses on more potentially suitable paths, and is demonstrated to effectively achieve a fairly high rank correlation of candidate architectures compared with RS.

The second phase involves the selection of the most promising architecture from the trained supernet, which is the primary purpose of most NAS tasks. Both SMASH \cite{smash_onehot} and \cite{under_oneshot} randomly selected a set of architectures from the supernet, and ranked them according to their performance. SMASH can obtain the validation performance of all selected architectures at the cost of a single training run for each architecture, as these architectures are assigned the weights generated by the hypernetwork. Besides, the authors in \cite{under_oneshot} observed that the architectures with a smaller symmetrized KL divergence value are more likely to perform better. This can be expressed as follows:

\begin{equation}
\begin{split}
    D_{\mathrm{SKL}} = D_{\mathrm{KL}}(p \| q)+D_{\mathrm{KL}}(q \| p) \\
   s.t. \,\, D_{\mathrm{KL}}(p \| q) =\sum_{i=1}^{n} p_{i} \log \frac{p_{i}}{q_{i}}
\end{split}
\end{equation}

\noindent where $(p_1,...,p_n)$ and $(q_1,...,q_n)$ indicate the predictions of the sampled architecture and one-shot model, respectively, and $n$ indicates the number of classes. The cost of calculating the KL value is very small; in \cite{under_oneshot}, only 64 random training data examples were used. Meanwhile, EA is also a promising search solution \cite{singlepath_nas,you2020greedynas}. For instance, SPOS \cite{singlepath_nas} uses EA to search for architectures from the supernet. It is more efficient than the EA methods introduced in Section \ref{section:model_generation}, because each sampled architecture only performs inference. The self-evaluated template network (SETN) \cite{oneshot_self} proposes an \textit{estimator} to predict the probability of each architecture having a lower validation loss. The experimental results show that SETN can potentially find an architecture with better performance than RS-based methods \cite{under_oneshot,smash_onehot}.

\subsection{Joint Hyperparameter and Architecture Optimization}

Most NAS methods fix the same setting of training-related hyperparameters during the whole search stage. After the search, the hyperparameters of the best-performing architecture are further optimized. However, this paradigm may result in suboptimal results as different architectures tend to fit different hyperparameters, making the model ranking unfair \cite{dong2020autohas}. Therefore, a promising solution is the joint \textit{hyperparameter and architecture optimization (HAO)} \cite{towards_nas_hpo,tabular_hao,dong2020autohas,dai2020fbnetv3}. We summary the existing joint HAO methods as follows.

Zela et al. \cite{towards_nas_hpo} cast NAS as a hyperparameter optimization problem, where the search spaces of NAS and standard hyperparameters are combined. They applied BOHB \cite{bohb}, an efficient HPO method, to optimize the architecture and hyperparameters jointly. Similarly, Dong et al. \cite{dong2020autohas} proposed a differentiable method, namely AutoHAS, which builds a Cartesian product of the search spaces of both NAS and HPO by unifying the representation of all candidate choices for the architecture (e.g., number of layers) and hyperparameters (e.g., learning rate). However, a challenge here is that the candidate choices for the architecture search space are usually categorical, while hyperparameters choices can be categorical (e.g., the type of optimizer) and continuous (e.g., learning rate). To overcome this challenge, AutoHAS discretizes the continuous
hyperparameters into a linear combination of multiple categorical bases. For example, the categorical bases for the learning rate are $\{0.1,0.2,0.3\}$, and then, the final learning rate is defined as $lr=w_1\times 0.1+w_2\times 0.2+w_3\times 0.3$. Meanwhile, FBNetv3 \cite{dai2020fbnetv3} jointly searches both architectures and the corresponding training recipes (i.e., hyperparameters). The architectures are represented with one-hot categorical variables and integral (min-max normalized) range variables, and the representation is fed to an encoder network to generate the architecture embedding. Then, the concatenation of architecture embedding and the training hyperparameters is used to train the accuracy predictor, which will be applied to search for promising architectures and hyperparameters at a later stage.

\subsection{Resource-aware NAS}

Early NAS studies \cite{nas_rl_zoph16,nasnet_zoph17,amoebanet} pay more attention to searching for neural architectures that achieve higher performance (e.g., classification accuracy), regardless of the associated resource consumption (i.e., the number of GPUs and time required). Therefore, many follow-up studies investigate resource-aware algorithms to trade off performance against the resource budget. To do so, these algorithms add computational cost to the loss function as a resource constraint. These algorithms differ in the type of computational cost, which may be 1) the parameter size; 2) the number of Multiply-ACcumulate (MAC) operations; 3) the number of float-point operations (FLOPs); or 4) the real latency. For example, MONAS \cite{monas} considers MAC as the constraint, and as MONAS uses a policy-based reinforcement-learning algorithm to search, the constraint can be directly added to the reward function. MnasNet \cite{mnasnet} proposes a customized weighted product to approximate a Pareto optimal solution:

\begin{equation}
\underset{m}{\operatorname{maximize}} \quad A C C(m) \times\left[\frac{L A T(m)}{T}\right]^{w}
\end{equation}

\noindent where $LAT(m)$ denotes measured inference latency of the model $m$ on the target device, $T$ is the target latency, and $w$ is the weight variable defined as:

\begin{equation}
w=\left\{\begin{array}{ll}
\alpha, & \text { if } LAT(m) \leq T \\
\beta, & \text { otherwise }
\end{array}\right.
\end{equation}

\noindent where the recommended value for both $\alpha$ and $\beta$ is $-0.07$.

In terms of a differentiable neural architecture search (DNAS) framework, the constraint (i.e., loss function) should be differentiable. For this purpose, FBNet \cite{fbnet} uses a latency lookup table model to estimate the overall latency of a network based on the runtime of each operator. The loss function is defined as

\begin{equation}
\mathcal{L}\left(a, \theta_{a}\right)=\operatorname{CE}\left(a, \theta_{a}\right) \cdot \alpha \log (\operatorname{LAT}(a))^{\beta}
\end{equation}

\noindent where $CE(a,\theta_a)$ indicates the cross-entropy loss of architecture $a$ with weights $\theta_a$. Similar to MnasNet \cite{mnasnet}, this loss function also comprises two hyperparameters that need to be set manually: $\alpha$ and $\beta$ control the magnitude of the loss function and the latency term, respectively. In SNAS \cite{snas}, the cost of time for the generated child network is linear to the one-hot random variables, such that the resource constraint's differentiability is ensured.

\section{Open Problems and Future Directions}\label{section:open_problems}

\begin{table*}[!ht]
    \centering
    \begin{center}
        
    \begin{tabular}{c|c|c}
    \hline
        \textbf{Category}&\textbf{Application}&\textbf{References}\\
        \hline

        \multirow{8}{*}{\makecell{Computer Vision\\(CV)}}
        & Medical Image Recognition & \cite{covid_benchmarking,automl_med_cls,automl_covidct} \\\cline{2-3}
        & Object Detection & \makecell{\cite{nasfpn,autofpn,efficientdet,chen2019detnas,guo2020hit,jiang2020sp}} \\\cline{2-3}
        & \makecell{Semantic Segmentation}  & \makecell{\cite{nasunet,auto_deeplab,nekrasov2019fast,bae2019resource,yang2019searching,dong2019neural,kim2019scalable}}\\\cline{2-3}
        & Person Re-identification & \makecell{\cite{quan2019auto}}\\\cline{2-3}
        & Super-Resolution & \cite{efficient_sr,chu2019fast,guo2020hierarchical} \\\cline{2-3}
        & Image Restoration & \cite{irnas} \\\cline{2-3}
        & \makecell{Generative Adversarial Network (GAN)} & \makecell{\cite{autogan,fu2020autogan,li2020gan,gao2020adversarialnas}}\\\cline{2-3}
        & Disparity Estimation & \cite{saikia2019autodispnet} \\\cline{2-3}
        & Video Task & \cite{peng2019video,ryoo2019assemblenet,nekrasov2020architecture,piergiovanni2019evolving}\\
        \hline

        \multirow{6}{*}{\makecell{Natural Language Processing\\(NLP)}} & Translation & \cite{fan2020searching} \\ \cline{2-3}
        & Language Modeling & \cite{jiang2019improved} \\\cline{2-3}
        & Entity Recognition &\cite{jiang2019improved} \\\cline{2-3}
        & Text Classification& \cite{chen2018exploring}\\\cline{2-3}
        & Sequential Labeling&\cite{chen2018exploring}\\\cline{2-3}
        & Keyword Spotting & \cite{mazzawi2019improving} \\
        \hline
        
        \multirow{8}{*}{Others} & Network Compression & \makecell{\cite{amc,autoprune,efficientprune,wang2020apq,dong2019network,huang2018learning,he2019meta,chin2018layercompensated}}\\\cline{2-3}
        & \makecell{Graph Neural Network (GNN)} &  \makecell{\cite{autognn}}\\ \cline{2-3}
        & Federate Learning & \makecell{\cite{he2020fednas,zhu2020real}}\\\cline{2-3}
        & Loss Function Search  & \makecell{\cite{li2019lfs,ru2020revisiting}}\\\cline{2-3}
        & Activation Function Search  & \makecell{\cite{ramach2017searching}}\\\cline{2-3}
        & Image Caption & \cite{wang2020evolutionary,wang2018neural}\\\cline{2-3}
        & Text to Speech (TTS) & \cite{semi_NAS} \\\cline{2-3}
        & Recommendation System&\cite{zhao2020amer,zhao2020autoemb,cheng2020differentiable}\\
        \hline
    \end{tabular}
    \end{center}
    \caption{Summary of the existing automated machine learning applications.}
    \label{table:nas_apps}
\end{table*}


This section discusses several open problems of the existing AutoML methods and proposes some future research directions.

\subsection{Flexible Search Space}

As summarized in Section \ref{section:model_generation}, there are various search spaces where the primitive operations can be roughly classified into pooling and convolution. Some spaces even use a more complex module (e.g., MBConv \cite{mnasnet}) as the primitive operation. Although these search spaces have been proven effective for generating well-performing neural architectures, all of them are based on human knowledge and experience, which inevitably introduce human bias, and hence, still do not break away from the human design paradigm. AutoML-Zero \cite{real2020automlzero} uses very simple mathematical operations (e.g., $\cos,\sin,\text{mean,std}$) as the primitive operations of the search space to minimize the human bias, and applies EA to discover complete machine learning algorithms. AutoML-Zero successfully designs two-layer neural networks based on these basic mathematical operations. Although the network searched by AutoML-Zero is much simpler than both human-designed and NAS-designed networks, the experimental results show the potential to discover a new model design paradigm with minimal human design. Therefore, the design of a more general, flexible, and free of human bias search space and the discovery of novel neural architectures based on this search space would be challenging and advantageous.

\subsection{Exploring More Areas}

As described in Section \ref{section:nas_performance}, the models designed by NAS algorithms have achieved comparable results in image classification tasks (CIFAR-10 and ImageNet) to those of manually designed models. Additionally, many recent studies have applied NAS to other CV tasks (Table \ref{table:nas_apps}). 

\begin{figure}[ht]
    \centering
    \includegraphics[width=0.48\textwidth]{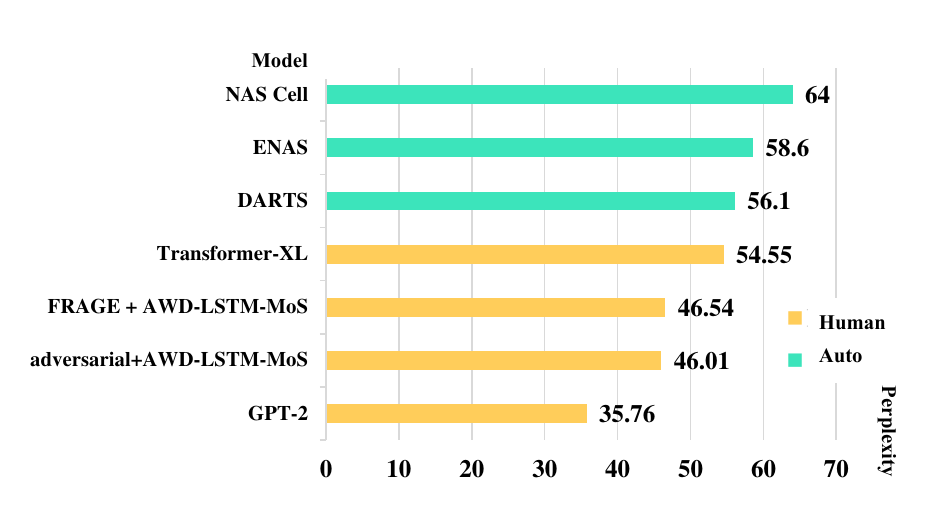}
    \caption{State-of-the-art models on the PTB dataset. The lower the perplexity, the better is the performance. The green bar represents the automatically generated model, and the yellow bar represents the model designed by human experts. Best viewed in color.}
    \label{fig:auto_vs_manual}
\end{figure}

However, in terms of the NLP task, most NAS studies have only conducted experiments on the PTB dataset. Besides, some NAS studies have attempted to apply NAS to other NLP tasks (shown in Table \ref{table:nas_apps}). However, Figure \ref{fig:auto_vs_manual} shows that, even on the PTB dataset, there is still a big gap in performance between the NAS-designed models (\cite{enas,darts_liu18,nas_rl_zoph16}) and human-designed models (GPT-2 \cite{GPT}, FRAGE AWD-LSTM-Mos \cite{frage}, adversarial AWD-LSTM-Mos \cite{adversarial_awd} and Transformer-XL \cite{transformerxl}). Therefore, the NAS community still has a long way to achieve comparable results to those of the models designed by experts on NLP tasks.

Besides the CV and NLP tasks, Table \ref{table:nas_apps} also shows that AutoML technique has been applied to other tasks, such as network compression, federate learning, image caption, recommendation system, and searching for loss and activation functions. Therefore, these interesting studies have indicated the potential of AutoML to be applied in more areas.


\subsection{Interpretability}

Although AutoML algorithms can find promising configuration settings more efficiently than humans, there is a lack of scientific evidence for illustrating why the found settings perform better. For example, in BlockQNN \cite{blockqnn}, it is unclear why the NAS algorithm tends to select the concatenation operation to process the output of each block in the cell, instead of the element-wise addition operation. Some recent studies \cite{under_oneshot,under_das,wong2016understanding} have shown that the explanation for these occurrences is usually hindsight and lacks rigorous mathematical proof. Therefore, increasing the mathematical interpretability of AutoML is an important future research direction.

\subsection{Reproducibility}

A major challenge with ML is reproducibility. AutoML is no exception, especially for NAS, because most of the existing NAS algorithms still have many parameters that need to be set manually at the implementation level; however, the original papers do not cover much detail. For instance, Yang et al. \cite{nas_eval} experimentally demonstrated that the seed plays an important role in NAS experiments; however, most NAS studies do not mention the seed set in the experiments. Besides, 
considerable resource consumption is another obstacle to reproduction. In this context, several NAS-Bench datasets have been proposed, such as NAS-Bench-101 \cite{ying2019nasbench}, NAS-Bench-201 \cite{nasbench201}, and NAS-Bench-NLP \cite{nasbenchnlp}. These datasets allow NAS researchers to focus on the design of optimization algorithms without wasting much time on the model evaluation.

\subsection{Robustness}

NAS has been proven effective in searching promising architectures on many open datasets (e.g., CIFAR-10 and ImageNet). These datasets are generally used for research; therefore, most of the images are well-labeled. However, in real-world situations, the data inevitably contain noise (e.g., mislabeling and inadequate information). Even worse, the data might be modified to be adversarial with carefully designed noises. Deep learning models can be easily fooled by adversarial data, and so can NAS. 

So far, there are a few studies \cite{is_nas_robust,NAS_Meets_Robustness,Chen2020OnRO,vargas2019evolving} have attempted to boost the robustness of NAS against adversarial data. Guo et al. \cite{NAS_Meets_Robustness} experimentally explored the intrinsic impact of network architectures on network robustness against adversarial attacks, and observed that densely connected architectures tend to be more robust. They also found that the flow of solution procedure (FSP) matrix \cite{FSP_matrix} is a good indicator of network robustness, i.e., the lower is the FSP matrix loss, the more robust is the network. Chen et al. \cite{Chen2020OnRO} proposed a robust loss function for effectively alleviating the performance degradation under symmetric label noise. The authors in \cite{vargas2019evolving} adopted EA to search for robust architectures from a well-designed and vast search space, where various adversarial attacks are used as the fitness function for evaluating the robustness of neural architectures.

\subsection{Joint Hyperparameter and Architecture Optimization}

Most NAS studies have considered HPO and AO as two separate processes. However, as already noted in Section \ref{section:model_generation}, there is a tremendous overlap between the methods used in HPO and AO, e.g., both of them apply RS, BO, and GO methods. In other words, it is feasible to jointly optimize both hyperparameters and architectures, which is experimentally confirmed by several studies \cite{tabular_hao,dong2020autohas,dai2020fbnetv3}. Thus, how to solve the problem of joint hyperparameter and architecture optimization (HAO) elegantly is a worthy studying issue.

\subsection{Complete AutoML Pipeline}

So far, many AutoML pipeline libraries have been proposed, but most of them only focus on some parts of the AutoML pipeline (Figure \ref{fig:pipeline_details}). For instance, TPOT \cite{Olson2016EvoBio}, Auto-WEAK \cite{auto_weak}, and Auto-Sklearn \cite{eff_automl_nips15} are built on top of scikit-learn \cite{sklearn} for building classification and regression pipelines, but they only search for the traditional ML models (such as SVM and KNN). Although TPOT involves neural networks (using Pytorch \cite{PyTorch} backend), it only supports an MLP network. Besides, Auto-Keras \cite{autokeras} is an open-source library developed based on Keras \cite{Keras}, which focuses more on searching for deep learning models and supports multi-modal and multi-task. NNI \cite{nni} is a more powerful and lightweight toolkit of AutoML, as its built-in capability contains automated feature engineering, hyperparameter optimization, and neural architecture search. Additionally, the NAS module in NNI supports both Pytorch \cite{PyTorch} and Tensorflow \cite{abadi2016tensorflow} and reproduces many SOTA NAS methods \cite{enas,darts_liu18,proxylessnas,pdarts,singlepath_nas,nas_random,ying2019nasbench}, which is very friendly for NAS researchers and developers. Besides, NNI also integrates scikit-learn features \cite{sklearn}, which is one step closer to achieving a complete pipeline. Similarly, Vega \cite{vega} is another AutoML algorithm tool that constructs a complete pipeline covering a set of highly decoupled functions: data augmentation, HPO, NAS, model compression, and full training. In summary, designing an easy-to-use and complete AutoML pipeline system is a promising research direction.
 



\subsection{Lifelong Learning}

Finally, most AutoML algorithms focus only on solving a specific task on some fixed datasets, e.g., image classification on CIFAR-10 and ImageNet. However, a high-quality AutoML system should have the capability of lifelong learning, i.e., it should be able to 1) efficiently \textbf{learn new data} and 2) \textbf{remember old knowledge}.


\subsubsection{Learn New Data}

First, the system should be able to reuse prior knowledge to solve new tasks (i.e., learning to learn). For example, a child can quickly identify tigers, rabbits, and elephants after seeing several pictures of these animals. However, the current DL models must be trained on considerable data before they can correctly identify images. A hot topic in this area is meta-learning, which aims to design models for new tasks using previous experience.

\textbf{Meta-learning.} Most of the existing NAS methods can search a well-performing architecture for a single task. However, they have to search for a new architecture on a new task; otherwise, the old architecture might not be optimal. Several studies \cite{continual_nas,Auto-meta,transferNAS,meta_NAS_fewshot} have combined meta-learning and NAS to solve this problem. Recently, Lian et al. \cite{transferNAS} proposed a novel and meta-learning-based \textit{transferable neural architecture search} method to generate a meta-architecture, which can adapt to new tasks easily and quickly through a few gradient steps. Another challenge of learning new data is few-shot learning scenarios, where there are only limited data for the new tasks. To overcome this challenge, the authors in \cite{Auto-meta} and \cite{continual_nas} applied NAS to few-shot learning, where they only searched for the most promising architecture and optimized it to work on multiple few-shot learning tasks. Elsken et al. \cite{meta_NAS_fewshot} proposed a gradient-based meta-learning NAS method, namely METANAS, which can generate \textit{task-specific} architectures more efficiently as it does not require meta-retraining.

\textbf{Unsupervised learning.} Meta-learning-based NAS methods focus more on labeled data, while in some cases, only a portion of the data may have labels or even none at all. Liu et al. \cite{UnNAS} proposed a general problem setup, namely \textit{unsupervised neural architecture search (UnNAS)}, to explore whether labels are necessary for NAS. They experimentally demonstrated that the architectures searched without labels are competitive with those searched with labels; therefore, labels are not necessary for NAS, which has provoked some reflection among researchers about which factors do affect NAS.


\subsubsection{Remember Old Knowledge}

An AutoML system must be able to constantly learn from new data, without forgetting the knowledge from old data. However, when we use new datasets to train a pretrained model, the model’s performance on the previous datasets is substantially reduced. Incremental learning can alleviate this problem. For example, Li and Hoiem \cite{lwf} proposed the \textit{learning without forgetting} (LwF) method, which trains a model using only new data while preserving its original capabilities. In addition, iCaRL \cite{icarl} makes progress based on LwF. It only uses a small proportion of old data for pretraining, and then gradually increases the proportion of a new class of data used to train the model.


\section{Conclusions}\label{section:conclusion}

This paper provides a detailed and systematic review of AutoML studies according to the DL pipeline (Figure \ref{fig:pipeline_details}), ranging from data preparation to model evaluation. Additionally, we compare the performance and efficiency of existing NAS algorithms on the CIFAR-10 and ImageNet datasets, and provide an in-depth discussion of different research directions on NAS: one/two-stage NAS, one-shot NAS, and joint HAO. We also describe several interesting open problems and discuss some important future research directions. Although research on AutoML is in its infancy, we believe that future researchers will effectively solve these problems. In this context, this review provides a comprehensive and clear understanding of AutoML for the benefit of those new to this area, and will thus assist with their future research endeavors.

\bibliographystyle{elsarticle-num} 
\bibliography{reference.bib}

\end{document}